\documentclass[10pt,twocolumn,letterpaper]{article}

\usepackage[accsupp]{axessibility}

\usepackage[pagenumbers]{cvpr} 

\usepackage[dvipsnames]{xcolor}

\usepackage{graphicx}
\usepackage{multirow}
\usepackage{mathtools}

\usepackage{pifont}
\usepackage{pdfrender}
\newcommand{\xmark}{\ding{55}}%
\usepackage{colortbl}

\newcommand{\stdv}[1]{\scriptsize$\pm$#1}

\newcommand*{\boldcheckmark}{%
  \textpdfrender{
    TextRenderingMode=FillStroke,
    LineWidth=.5pt, 
  }{\checkmark}%
}
\newcommand*{\boldxmark}{%
  \textpdfrender{
    TextRenderingMode=FillStroke,
    LineWidth=.5pt, 
  }{\xmark}%
}
\definecolor{Gray}{gray}{0.95}

\usepackage{amsmath}
\usepackage{amssymb}
\usepackage{graphicx}
\usepackage{algorithm} 
\usepackage{algpseudocode}

\definecolor{blackpink}{rgb}{0.6,0,0.6}

\usepackage{lipsum}

\newcommand\blfootnote[1]{%
  \begingroup
  \renewcommand\thefootnote{}\footnote{#1}%
  \addtocounter{footnote}{-1}%
  \endgroup
}
\newcommand*{\affmark}[1][*]{\textsuperscript{#1}}

\definecolor{cvprblue}{rgb}{0.21,0.49,0.74}
\definecolor{Gray}{gray}{0.95}
\usepackage[pagebackref,breaklinks,colorlinks,citecolor=cvprblue]{hyperref}

\def\paperID{12211} 
\def\confName{CVPR}
\def\confYear{2024}

\title{Enhancing Intrinsic Features for Debiasing via Investigating \\ Class-Discerning Common Attributes in Bias-Contrastive Pair}

\author{%
  Jeonghoon Park\affmark[*1], Chaeyeon Chung\affmark[*1], Juyoung Lee\affmark[2], Jaegul Choo\affmark[1]\\
  \affmark[1]Korea Advanced Institute of Science and Technology, South Korea, \affmark[2]Kakao Corp., South Korea.\\
  \texttt{\footnotesize \affmark[1]\{jeonghoon\_park, cy\_chung, jchoo\}@kaist.ac.kr}, \texttt{\footnotesize \affmark[2]michael.l22@kakaocorp.com} \\
}

\begin{document}
\maketitle
\begin{abstract}
In the image classification task, deep neural networks frequently rely on bias attributes that are spuriously correlated with a target class in the presence of dataset bias, resulting in degraded performance when applied to data without bias attributes.
The task of debiasing aims to compel classifiers to learn intrinsic attributes that inherently define a target class rather than focusing on bias attributes.
While recent approaches mainly focus on emphasizing the learning of data samples without bias attributes (i.e., bias-conflicting samples) compared to samples with bias attributes (i.e., bias-aligned samples),  they fall short of directly guiding models where to focus for learning intrinsic features.
To address this limitation, this paper proposes a method that provides the model with explicit spatial guidance that indicates the region of intrinsic features. 
We first identify the intrinsic features by investigating the class-discerning common features between a bias-aligned (BA) sample and a bias-conflicting (BC) sample (i.e., bias-contrastive pair).
Next, we enhance the intrinsic features in the BA sample that are relatively under-exploited for prediction compared to the BC sample. 
To construct the bias-contrastive pair without using bias information, we introduce a bias-negative score that distinguishes BC samples from BA samples employing a biased model.
The experiments demonstrate that our method achieves state-of-the-art performance on synthetic and real-world datasets with various levels of bias severity.
\end{abstract}

\blfootnote{* indicates equal contribution.}
\vspace{-4mm}

\section{Introduction}
Deep neural networks in image classification~\cite{szegedy2015going, He2015resnet, vgg, wideresnet} are known to be vulnerable to the dataset bias~\cite{unbiaslook2011torralba}, which refers to a spurious correlation between the target classes and the peripheral attributes.
Basically, image classification aims to learn intrinsic attributes — the visual features that inherently define a target class — that generally appear across the samples in the class.
However, when the dataset bias exists in the training data, the models tend to use the frequently appearing peripheral attribute (\ie, bias attribute) to predict the class unintentionally.
For instance, if airplanes in the training images are mostly in the sky, a model can heavily rely on the sky to predict an image as an airplane class due to its high correlation with the airplane class.
This indicates that the model is biased towards the bias attribute (\eg, sky) rather than focusing on intrinsic features (\eg, the shape of wings or the body) when making decisions.
As a result, even though the biased model achieves high accuracy on the samples including bias attributes (\eg, airplanes in the sky), termed as bias-aligned (BA) samples, it may fail to accurately predict samples devoid of such bias attributes (\eg, airplanes on the runway), referred to as bias-conflicting (BC) samples.

In this regard, debiasing aims to encourage the model to focus on intrinsic attributes rather than bias attributes when dataset bias exists.
One straightforward approach is utilizing prior knowledge regarding bias (\eg, labels for bias attribute) to inform the model which attributes to focus on or not to focus on~\cite{kim2019LNL,wang2018hex,bahng2019rebias,EnD}.
However, acquiring such bias information is often infeasible in real-world scenarios.
Therefore, recent studies~\cite{nam2020learning, liu2021just,lee2022revisiting, disentangled, hwangselecmix} have proposed debiasing methods that do not require bias information.
They identify and emphasize BC samples during the training using an additional biased classifier that mainly learns the bias attributes.
However, such a training strategy fails to directly indicate where the model should focus to learn the intrinsic features.

To address this issue, we present a debiasing approach that explicitly informs the model of the region of the intrinsic features during the training while not using bias labels.
While the intrinsic features in the unbiased dataset can simply be identified in generally appearing features in the training samples, generally appearing features in the biased dataset inevitably include bias features.
Therefore, we identify the intrinsic features in the biased dataset by investigating the common features between a BA and a BC sample (i.e., a bias-contrastive pair).
Here, the common features also need to be class-discerning since the common features might include irrelevant environmental features.
For example, in the above scenario, the common feature between an airplane in the sky (BA sample) and an airplane on the runway (BC sample) might include the features of wings, the body, and trees.
In this case, the intrinsic features are the shape of the wings and the body that can distinguish the airplane class from the others.

Specifically, we introduce an intrinsic feature enhancement (IE) weight that identifies the spatial regions of intrinsic features commonly appearing in a bias-contrastive pair.
We leverage an auxiliary sample in addition to the original input to construct the bias-contrastive pair. 
Since the majority of the original input from training samples are BA samples, we mainly adopt the BC samples as the auxiliary sample.
To achieve this without bias information, we present a bias-negative (BN) score that identifies BC samples by employing a classification loss of a biased model.
Our IE weight investigates common features in the bias-contrastive pair and identifies the class-discerning features among the common features.
Within the identified intrinsic features, we enhance the features that are relatively under-exploited in the BA samples compared to the BC samples.
In this way, we can explicitly provide our model with spatial guidance for intrinsic attributes while not using bias labels.

We verify the effectiveness of our method on both synthetic and real-world datasets with various levels of bias severity.
Furthermore, the in-depth analysis demonstrates that our method successfully guides the model to make predictions based on the intrinsic features.

\section{Related work}
\label{sec:relatedwork}
\noindent\textbf{Debiasing with bias information.\enskip}
Previous approaches~\cite{kim2019LNL, EnD, sagawa2019distributionally, wang2018hex, geirhos2018imagenettrained, bahng2019rebias} utilize bias labels or predefined bias types to encourage the model to learn intrinsic attributes for debiasing.
Kim \etal~\cite{kim2019LNL}, Tartaglione \etal~\cite{EnD}, and Sagawa \etal~\cite{sagawa2019distributionally} employ bias labels to encourage the model not to learn specific bias features.
Wang \etal~\cite{wang2018hex} and Bahng \etal~\cite{bahng2019rebias} predefine the bias type (e.g., color, texture, etc.) and utilize such prior knowledge to supervise models to be robust against such predefined bias type.
However, obtaining bias information requires additional cost, which is often infeasible in the real world.

\noindent\textbf{Debiasing without bias information.\enskip}
Recent studies~\cite{darlow2020latent,huangRSC2020,nam2020learning, disentangled, lee2022revisiting, liu2021just, hwangselecmix, masktune2022, Zhang_2023_CVPR, dcwp2023park,kwon_dash_2022} propose debiasing strategies that do not require bias information.
Nam \etal~\cite{nam2020learning} present an approach that encourages the model to concentrate on BC samples during the training process considering that the bias attributes are easier to learn than intrinsic attributes.
Instead of using bias information, they additionally train a biased model that mainly learns bias attributes and regard the samples that are not easily trained by the biased model as BC samples.
Lee \etal~\cite{lee2022revisiting} reveal that BC samples serve as noisy samples when training the additional biased model and propose a method to eliminate such BC samples using multiple biased models.
Liu \etal~\cite{liu2021just} regard the samples misclassified by the model trained with empirical risk minimization as BC samples and emphasize them during training of a debiased model.
Also, MaskTune~\cite{masktune2022} expects the model to learn intrinsic features by fine-tuning the model with the data whose already-explored area is masked out using Grad-CAM~\cite{gradcam}.

Another stream of approaches~\cite{biaswap,disentangled, hwangselecmix} synthesize samples having similar characteristics with BC samples and employ them to train a debiased model.
Kim \etal~\cite{biaswap} synthesize images without bias attributes leveraging an image-to-image translation model~\cite{park2020swapping}.
Lee \etal~\cite{disentangled} and Hwang \etal~\cite{hwangselecmix} augment BC samples in the feature space by employing the disentangled representations and mixup~\cite{zhang2018mixup}, respectively.
A recent pair-wise debiasing method $\mathcal{X}^2$-model~\cite{Zhang_2023_CVPR} encourages the model to retain intra-class compactness using samples generated via feature-level interpolation between BC and BA samples.
However, such approaches lack explicit supervision about which features to focus on to learn intrinsic features.
To address this issue, we present a debiasing method that provides spatial guidance to encourage a model to learn intrinsic features during the training while not using bias labels.
We design our model architecture using bias-contrastive pairs referring to the previous studies~\cite{hwangselecmix, Zhang_2023_CVPR}.

\begin{figure*}[t]
\centering
     \includegraphics[width=1.0\textwidth]{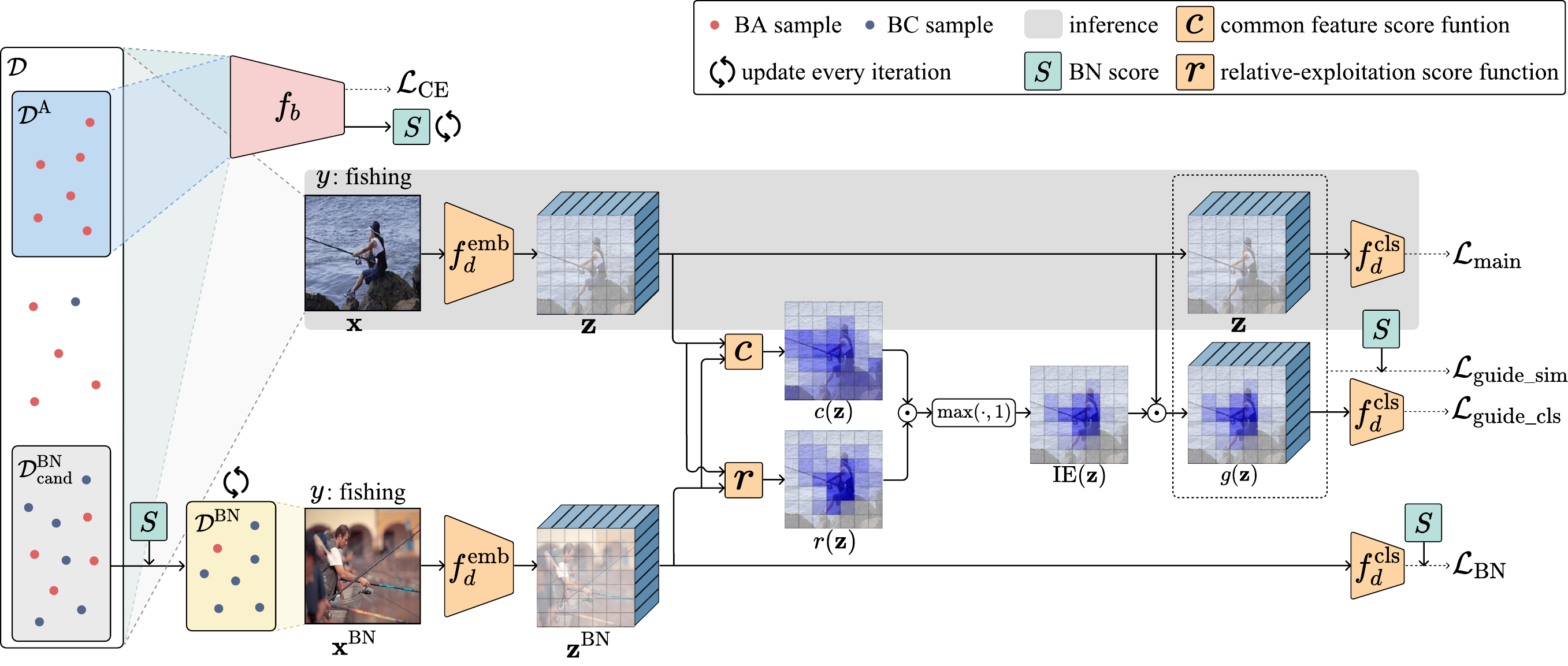}
     \caption{Overview of our method. 
     We provide explicit spatial guidance $g(\mathbf{z})$ for a debiased model $f_d$, which is described with $f_d^\text{emb}$ and $f_d^\text{cls}$, to learn intrinsic features.
     To achieve this, we leverage a bias-contrastive pair, $\mathbf{x}$ and $\mathbf{x}^\text{BN}$ from the same target class $y$.
     $g(\mathbf{z})$ highlights intrinsic features that are relatively under-exploited in $\mathbf{z}$ compared to $\mathbf{z}^\text{BN}$, calculated by common feature score $c$ and relative-exploitation score $r$.
     Here, we mainly adopt BC samples from $\mathcal{D}^\text{BN}_\text{cand}$ to construct $\mathcal{D}^\text{BN}$, where we sample $\mathbf{x}^\text{BN}$.
     $\mathcal{D}^\text{BN}$ is updated every iteration using the BN score $S$, which is also updated every iteration.
     At the inference, we only use $f_d$ in the gray-colored area. 
     }
     \vspace{-2mm}
     \label{fig:overview}
\end{figure*}

\section{Methodology}

\subsection{Overview}
As shown in Fig.~\ref{fig:overview}, our framework consists of a biased model $f_b$ that focuses on bias attributes and a debiased model $f_d$ that learns debiased representations. 
We use BiasEnsemble (BE)~\cite{lee2022revisiting} as a backbone, where $f_b$ is trained with bias-amplified dataset $\mathcal{D}^\text{A}$ which mainly consists of BA samples, while $f_d$ concentrates on the samples that $f_b$ fails to learn.
Our method provides $f_d$ with spatial guidance for intrinsic features using a bias-contrastive pair: an input $\mathbf{x}$ and an auxiliary input $\mathbf{x}^\text{BN}$.
We denote the auxiliary input $\mathbf{x}^\text{BN}$ as a bias-negative (BN) sample because we primarily adopt samples devoid of bias attributes.
We sample an image $\mathbf{x}$ from the original training data $\mathcal{D}$, and $\mathbf{x}^\text{BN}$ from a BN dataset $\mathcal{D}^\text{BN}$ which mainly consists of BC samples.
$\mathcal{D}^\text{BN}$ is updated every iteration to mainly include BC samples using the BN score $S$ that employs $f_b$ to identify BC samples.
The BN score is also updated every iteration.
Given the intermediate features $\mathbf{z}$ and $\mathbf{z}^\text{BN}$, we first extract the common features between the bias-contrastive pair ($c(\mathbf{z})$ in Fig.~\ref{fig:overview}).
Also, we identify the class-discerning features that are relatively under-exploited in $\mathbf{z}$ compared to $\mathbf{z}^\text{BN}$ ($r(\mathbf{z})$ in Fig.~\ref{fig:overview}).
Next, we calculate the IE weight that indicates relatively under-exploited intrinsic features in $\mathbf{z}$ based on $c(\mathbf{z})$ and $r(\mathbf{z})$ ($\text{IE}(\mathbf{z})$ in Fig.~\ref{fig:overview}).
Finally, we obtain the guidance $g(\mathbf{z})$ that emphasizes the region of intrinsic feature in $\mathbf{z}$ during the training.
At the inference, we utilize $f_d$ without $\mathbf{x}^\text{BN}$, as in a gray-colored area of Fig.~\ref{fig:overview}.

\subsection{Constructing bias-negative dataset}
\label{subsec:BNscore} 
We construct a BN dataset $\mathcal{D}^\text{BN}$, where we sample $\mathbf{x}^\text{BN}$ during the training.
As the majority of the training dataset is BA samples, we aim to mainly adopt BC samples as $\mathbf{x}^\text{BN}$ to construct bias-contrastive pairs.
To achieve this, we first construct $\mathcal{D}^\text{BN}_\text{cand}$, a candidate dataset for $\mathcal{D}^\text{BN}$, that contains roughly identified BC samples.
During the training, we dynamically update $\mathcal{D}^\text{BN}$ every iteration to mainly adopt BC samples from $\mathcal{D}^\text{BN}_\text{cand}$ using our newly proposed BN score.

\noindent\textbf{Constructing candidate dataset $\mathcal{D}^\text{BN}_\text{cand}$.\enskip}
To roughly identify BC samples in $\mathcal{D}$, we filter out easily learned BA samples from $\mathcal{D}$ using multiple biased models, following BE~\cite{lee2022revisiting}.
Since the bias features are easier to learn than the intrinsic features~\cite{nam2020learning}, each biased model is trained only for a few iterations so that BC samples can be distinguished from the easily learned BA samples.
Inspired by JTT~\cite{liu2021just}, we regard the samples that are incorrectly predicted by the majority of the biased models as BC samples.
Finally, we construct $\mathcal{D}^\text{BN}_\text{cand}$ with the roughly identified BC samples.

\noindent\textbf{Adopting BC samples with BN score.\enskip}
We introduce a BN score to update $\mathcal{D}^\text{BN}$ to primarily exploit BC samples as $\mathbf{x}^\text{BN}$ from $\mathcal{D}^\text{BN}_\text{cand}$ during training $f_d$.
Considering the unavailability of bias labels, the BN score employs $f_b$ to further exclude BA samples from $\mathcal{D}^\text{BN}_\text{cand}$.
As training proceeds, $f_b$ is overfitted to the bias attributes in $\mathcal{D}^\text{A}$ and outputs a high probability on the ground-truth label for the samples that have similar bias features with samples in $\mathcal{D}^\text{A}$.
This indicates that samples whose $f_b$ loss decreases as training proceeds are likely to have bias attributes learned from $\mathcal{D}^\text{A}$.
Such samples disturb the extraction of intrinsic features when selected as $\mathbf{x}^\text{BN}$.
To validate this, we investigate the samples in $\mathcal{D}^\text{BN}_\text{cand}$ whose $f_b$ loss at the later stage of training (50K-th iteration) decreases compared to the early stage of training (1K-th iteration).
The result shows that 95.63\% of them are BA samples.
We use the BFFHQ dataset~\cite{biaswap} with a bias severity of 1\% for the analysis. Further details of the dataset are described in Sec.~\ref{subsec:expsetting}.

In this regard, we design a BN score to exclude the samples with decreasing $f_b$ loss from the $\mathcal{D}^\text{BN}_\text{cand}$ to construct $\mathcal{D}^\text{BN}$ by tracking the $f_b$ loss during training $f_d$.
First, the $f_b$ loss of $\mathbf{x}$ at the $t$-th iteration is calculated as follows:
\begin{equation}
    l_{t}(\mathbf{x}) = \alpha_l \cdot  \mathcal{L}_\text{CE}(f_b(\mathbf{x}), y)+ (1-\alpha_l)\cdot l_{t-1}(\mathbf{x}),
\end{equation}
where $\mathcal{L}_\text{CE}(f_b(\mathbf{x}), y)$ indicates the cross-entropy (CE) loss of $\mathbf{x}$ on its ground-truth label $y$ and $\alpha_l$ is a hyperparameter for the exponential moving average (EMA).
We employ EMA to enable a stable tracking of the classification losses.
Note that $l_t$ is updated only for the samples in a mini-batch at the $t$-th iteration. 

The BN score tracks $l_{t}(\mathbf{x})$ compared to the loss recorded at the early stage of training. 
The BN score at the $t$-th iteration is formulated as follows:
\begin{equation}
    s_t(\mathbf{x}) = \alpha_s\cdot (l_{t}(\mathbf{x})-l_\text{ref}(\mathbf{x})) + (1-\alpha_s)\cdot s_{t-1}(\mathbf{x}),
\label{eq:bnscore}
\end{equation}
where $\alpha_s$ is a hyperparameter for the EMA and $l_\text{ref}(\mathbf{x})$ denotes the reference loss of $\mathbf{x}$ that is first recorded after a few iterations of training.
We exploit EMA to stabilize the tracking.
Note that we update $s_t$ only for the samples in a mini-batch at the $t$-th iteration. 
The negative value of $s_t(\mathbf{x})$ indicates that the loss of $\mathbf{x}$ decreased compared to the early stage of training, which means that the sample is likely to contain bias attributes.

\noindent\textbf{Updating $\mathcal{D}^\text{BN}$ with BN score.\enskip}
At every iteration, we update $\mathcal{D}^\text{BN}$ to exclude the newly detected BA samples whose BN score $s_t(\mathbf{x})$ is smaller than zero as follows:
\begin{equation}
    \mathcal{D}^\text{BN}_{t} = \{\mathbf{x} \mid s_t(\mathbf{x}) > 0, \mathbf{x} \sim \mathcal{D}^\text{BN}_\text{cand} \},
\end{equation}
where $\mathcal{D}^\text{BN}_{t}$ indicates $\mathcal{D}^\text{BN}$ at the $t$-th iteration.
We employ $\mathbf{x}^\text{BN}\sim\mathcal{D}^\text{BN}_t$ as an auxiliary input at the $t$-th iteration.
In this way, we can construct a bias-contrastive pair that encourages the intrinsic attributes to be extracted as their common features.
We abbreviate $\mathcal{D}^\text{BN}_{t}$ as $\mathcal{D}^\text{BN}$ for brevity in the rest of the paper.

\subsection{Intrinsic feature enhancement}
To emphasize the intrinsic features in $f_d$, we introduce the intrinsic feature enhancement (IE) weight that imposes a high value on the intrinsic features.
The IE weight identifies the region of intrinsic features from bias-contrastive pairs by investigating 1) their common features with common feature score $c$ and 2) class-discerning features that are relatively under-exploited in the input with relative-exploitation score $r$.
For the explanation, we split $f_d$ into two parts. $f_d^{\text{emb}} : \mathbb{R}^{H \times W \times 3} \to \mathbb{R}^{h \times w \times c}$ maps an input to the intermediate feature, and $f_d^{\text{cls}} : \mathbb{R}^{h \times w \times c} \to \mathbb{R}^{C}$ is composed of the average pooling and the linear classifier and outputs the classification logits, where $f_d(\mathbf{x}) = f_d^{\text{cls}}\left(f_d^{\text{emb}}(\mathbf{x})\right)$.

First, given the input $\mathbf{x}$, common feature score $c$ identifies the features that are similar to the features in $\mathbf{x}^\text{BN}$ that has the same class label as $\mathbf{x}$ while not having bias attributes. 
Specifically, we extract the intermediate features $\mathbf{z}=f_d^\text{emb}(\mathbf{x})$ and $\mathbf{z}^\text{BN}=f_d^\text{emb}(\mathbf{x}^\text{BN})$, respectively.
Next, we obtain the common feature score of $\mathbf{z}$ (\ie, $c(\mathbf{z}) \in \mathbb{R}^{h \times w}$).
Given the $n$-th feature of $\mathbf{z}$ (\ie, $\mathbf{z}_{n} \in \mathbb{R}^{c}$), let $i^*$-th feature of $\mathbf{z}^\text{BN}$ (\ie, $\mathbf{z}^\text{BN}_{i^*} \in \mathbb{R}^{c}$) be the most similar feature to $\mathbf{z}_n$, where $i^* = \underset{i}{\arg\max}\left(\mathbf{z}^\text{BN}_i\cdot\mathbf{z}_n\right)$.
Then, the $n$-th element of $c(\mathbf{z})$ denotes the similarity score between $\mathbf{z}_n$ and $\mathbf{z}^\text{BN}_{i^*}$, which is formulated as follows:
\begin{equation}
    c(\mathbf{z})_{n} =\frac{\mathbf{z}_{i^*}^\text{BN}\cdot\mathbf{z}_{n}}{\max_{i, j}(\mathbf{z}_i^\text{BN}\cdot\mathbf{z}_j)},
\end{equation}
where $\cdot$ indicates a dot product operation.
We adopt the dot product for the similarity metric to consider both the scale and the direction of the features. 
The max normalization is employed to limit the score to less than one.
We consider the features with a high common feature score $c$ in $\mathbf{z}$ as features that have a high likelihood of being intrinsic features.

Next, the relative-exploitation score $r$ identifies class-discerning features that are relatively under-exploited in $\mathbf{x}$ compared to $\mathbf{x}^\text{BN}$.
Since most of the $\mathbf{x}^\text{BN}$ does not contain bias attributes, we identify class-discerning intrinsic features by investigating the features that are mainly used to predict $\mathbf{x}^\text{BN}$ as its target label.
At the same time, we identify the features that are under-exploited in the $\mathbf{x}$ compared to the $\mathbf{x}^\text{BN}$.
To achieve this, we use a visual explanation map of Grad-CAM~\cite{gradcam} that imposes a higher value on the features that have more contribution to predicting a specific label.
We calculate the explanation map $\text{E}(\mathbf{z})$ and $\text{E}(\mathbf{z}^\text{BN})$ with respect to their ground-truth labels.
We apply max normalization to the explanation maps to compare the relative importance of the features in prediction.
We compare the $n$-th value of $\text{E}(\mathbf{z})$ (\ie, $\text{E}(\mathbf{z})_{n}$) with the $i^*$-th value of $\text{E}(\mathbf{z}^\text{BN})$, where $i^*$ is the index of the feature in $\mathbf{z}^\text{BN}$ that is the most similar with $\mathbf{z}_n$.
Accordingly, the $n$-th element of $r(\mathbf{z})\in \mathbb{R}^{h\times w}$ is calculated as: 
\begin{equation}
    r(\mathbf{z})_{n} = \left(\frac{2\text{E}(\mathbf{z}^\text{BN})_{i^*}}{\text{E}(\mathbf{z}^\text{BN})_{i^*}+\text{E}(\mathbf{z})_{n}}\right)^\tau,
\end{equation}
where $\tau$ is the amplification factor.
The score becomes larger than one when the $\mathbf{z}_n$ is relatively under-exploited than $\mathbf{z}^\text{BN}_{i^*}$ for prediction.
When $\mathbf{z}^\text{BN}_{i^*}$ is not used for discerning the class, $\text{E}(\mathbf{z}^\text{BN})_{i^*}$ becomes close to zero and the score converges to zero.

Finally, $n$-th element of the $\text{IE}(\mathbf{z})$ is defined as:
\begin{equation}
    \text{IE}(\mathbf{z})_{n} = \max(c(\mathbf{z})_{n} \odot r(\mathbf{z})_{n}, 1),
\end{equation}
where $\odot$ indicates the element-wise multiplication.
The IE weight has a large value on the features of $\mathbf{x}$ that commonly appear in $\mathbf{x}^\text{BN}$ but has not exploited enough for the prediction of $\mathbf{x}$.
We clip the values to be larger than one to enhance only the relatively under-exploited features in $\mathbf{z}$ while preserving the other features.

Using the IE weight, we obtain the guidance $g(\mathbf{z})$ that emphasizes the intrinsic features in $\mathbf{z}$ as follows:
\begin{equation}
    g(\mathbf{z}) = \mathbf{z}\odot \text{IE}(\mathbf{z}).
\end{equation}
We broadcast $\text{IE}(\mathbf{z})$ to match its shape with $\mathbf{z}$ before multiplication.
During the training, this spatial guidance informs the model of where to focus to learn intrinsic features from training samples.

\begin{algorithm}
	\caption{Debiasing with the intrinsic feature guidance}\label{alg1}
        \hspace*{\algorithmicindent} \textbf{Input:} pretrained biased models, 
        \\\hspace*{\algorithmicindent} training dataset $\mathcal{D}$, biased model $f_b$, debiased model 
        \\\hspace*{\algorithmicindent}$f_d$, reference iteration $T_1$ for BN score, starting
        \\\hspace*{\algorithmicindent}iteration $T_2(\geq T_1)$ to apply intrinsic feature guidance  \\
        \hspace*{\algorithmicindent} \textbf{Output:} trained debiased model $f_d$
        
	\begin{algorithmic}[1]
            \State Construct $\mathcal{D}^\text{A}, \mathcal{D}^\text{BN}_\text{cand}$ from $\mathcal{D}$ using pretrained biased models
            \For {every iteration $t$}
                \State Sample $\mathbf{x} \sim \mathcal{D}$  
                \If{$t\geq T_1$ \textbf{and} $l_\text{ref}(\mathbf{x})$ is not initialized}
                        \State $l_\text{ref}(\mathbf{x}) \gets l(\mathbf{x})$
                \EndIf
                
                \If{$\mathbf{x} \in \mathcal{D}^\text{A}$}
                        \State Train $f_b(\mathbf{x})$ with $\mathcal{L}_\text{CE}$ 
                \EndIf

                \If{$t <$ $T_2$}
                \Comment{Train w/o guidance}
                    \State Train $f_d(\mathbf{x})$ with $\mathcal{L}_\text{main}$
                \ElsIf{$t \geq$ $T_2$}
                \Comment{Train w/ guidance}
                    \State Update $\mathcal{D}^{\text{BN}}$
                    \State Sample $\mathbf{x}^\text{BN} \sim \mathcal{D}^\text{BN}$  
                    \State Train $f_d(\mathbf{x}, \mathbf{x}^\text{BN})$ with $\mathcal{L}_\text{total}$
                    
                \EndIf

                \EndFor
	\end{algorithmic} 
\end{algorithm} 

\subsection{Training with intrinsic feature guidance}
We basically train $f_d$ with the CE loss as follows:
\begin{equation}
    \mathcal{L}_\text{main} = w(\mathbf{x})\mathcal{L}_\text{CE}(f_d(\mathbf{x}), y) 
\end{equation}
where $w(\mathbf{x})$ is the sample reweighting value of $\mathbf{x}$~\cite{nam2020learning}.
$w(\mathbf{x})$ emphasizes the samples that $f_b$ fails to learn, which are mostly BC samples.
The detailed description of $w(\mathbf{x})$ is included in the Supplementary.

In addition, we guide the model to focus on the region of intrinsic features through a guidance loss and a BN loss.
We observe that the BN score has a higher value on the BC samples compared to the BA samples as training $f_b$ proceeds (See Sec.~\ref{subsec:analysis_bn}). 
In this respect, we employ the BN score of $\mathbf{x}^\text{BN}$ (\ie, $s(\mathbf{x}^\text{BN})$) to upweight the loss when BC samples are adopted as $\mathbf{x}^\text{BN}$.
Here, we clip the value of loss weight $s(\mathbf{x}^\text{BN})$ to be larger than zero.

\noindent\textbf{Guidance loss.\enskip}
To guide the model to exploit the intrinsic features from $\mathbf{x}$, we minimize the L1 distance between $g(\mathbf{z})$ and $\mathbf{z}$ as follows:
\begin{equation}
    \mathcal{L}_\text{guide\_sim} = s(\mathbf{x}^\text{BN})\lVert \text{GAP}(\mathbf{z})-\text{GAP}\left({g(\mathbf{z})}\right)\rVert_1,
\end{equation}
where GAP represents the global average pooling.
$s(\mathbf{x}^\text{BN})$ is multiplied as a loss weight to impose a high weight on the loss when BC samples are selected as $\mathbf{x}^\text{BN}$.

Also, we apply the CE loss to the guidance $g(\mathbf{z})$ to encourage it to include the intrinsic features that contribute to the correct prediction as follows:
\begin{equation}
    \mathcal{L}_\text{guide\_cls} = w(\mathbf{x})\mathcal{L}_\text{CE}\left(f_d^\text{cls}(g(\mathbf{z})), y\right),
\end{equation}
where $w(\mathbf{x})$ is the reweighting value as in $\mathcal{L}_\text{main}$.

Finally, our guidance loss $\mathcal{L}_\text{guide}$ is calculated as follows:
\begin{equation}
    \mathcal{L}_\text{guide}= \lambda_\text{sim}\mathcal{L}_\text{guide\_sim}+\mathcal{L}_\text{guide\_cls},
\end{equation}
where $\lambda_\text{sim}$ is a hyperparameter to control the relative significance between the losses. 
We set $\lambda_\text{sim}$ set as 0.1.

\noindent\textbf{BN loss.\enskip}
We also employ the CE loss on $\mathbf{x}^\text{BN}$ to encourage the model to learn class-discerning features.
This enables the IE weight to find intrinsic features among the common features.
The BN loss is defined as:
\begin{equation}
    \mathcal{L}_\text{BN} = s(\mathbf{x}^\text{BN})\mathcal{L}_\text{CE}(f_d(\mathbf{x}^\text{BN}), y). 
\end{equation}
Here, we also exploit $s(\mathbf{x}^\text{BN})$ to impose high weight on the loss when $\mathbf{x}^\text{BN}$ is a BC sample.

\noindent\textbf{Overall objective function.\enskip}
In summary, the overall objective function is defined as follows:
\begin{equation}
    \mathcal{L}_\text{total} =  \lambda_\text{main}\mathcal{L}_\text{main}+ \mathcal{L}_\text{guide}+\mathcal{L}_\text{BN},
    \label{eq:tot_loss}
\end{equation}
$\lambda_\text{main}$ is the constant value that linearly increases from zero to one during training $f_d$ with the guidance.
This prevents the model from focusing on bias features in $\mathbf{x}$ in the early phase.
The overall process of our method is provided in Algorithm~\ref{alg1}. 
Here, we set $T_1$ and $T_2$ as 1K and 10K, respectively.
Note that all the hyperparameters are identically applied across different datasets and bias severities.
We provide further details of the training and the implementation in the Supplementary.

\begin{table*}
\begin{center}
\resizebox{1.0\linewidth}{!}{
\setlength{\tabcolsep}{1.0em}
\def\arraystretch{1.0}%
\begin{tabular}{l|cccc|cccc|cc}
\toprule 
\multicolumn{1}{c}
{\multirow{2}{*}{Method}} &
 \multicolumn{4}{c}{Waterbirds} & \multicolumn{4}{c}{BFFHQ} & \multicolumn{2}{c}{BAR}
\\ \cmidrule(lr){2-5}\cmidrule(lr){6-9}\cmidrule(lr){10-11} 
\multicolumn{1}{c}{}&
 0.5 & 1.0 & 2.0 & 5.0 &  0.5 & 1.0 & 2.0 & 5.0 & 1.0 & 5.0 
\\
\midrule

Vanilla~\cite{He2015resnet}
& 57.41 & 58.07 & 61.04 & 64.13 & 55.64 & 60.96 & 69.00 & 82.88 & 70.55 & 82.53\\
HEX~\cite{wang2018hex}
& 57.88 & 58.28 & 61.02 & 64.32 & 56.96 & 62.32 & 70.72 & 83.40 & 70.48 & 81.20\\
LNL~\cite{kim2019LNL}
& 58.49 & 59.68 & 62.27 & 66.07 & 56.88 & 62.64 & 69.80 & 83.08 & - & -\\
EnD~\cite{EnD}
& 58.47 & 57.81 & 61.26 & 64.11 & 55.96 & 60.88 & 69.72 & 82.88 & - & -\\
ReBias~\cite{bahng2019rebias}
& 55.44 & 55.93 & 58.53 & 62.14 & 55.76 & 60.68 & 69.60 & 82.64 & 73.04 & 83.90\\
LfF~\cite{nam2020learning}
& 60.66 & 61.78 & 58.92 & 61.43 & 65.19 & 69.24 & 73.08 & 79.80 & 70.16 & 82.95\\
DisEnt~\cite{disentangled}
& 59.59 & 60.05 & 59.76 & 64.01 & 62.08 & 66.00 & 69.92 & 80.68 & 70.33 & 83.13\\
LfF+BE~\cite{lee2022revisiting}
& 61.22 & 62.58 & 63.00 & 63.48 & 67.36 & 75.08 & 80.32 & 85.48 & 73.36 & 83.87\\
DisEnt+BE~\cite{lee2022revisiting}
& 51.65 & 54.10 & 53.43 & 54.21 & 67.56 & 73.48 & 79.48 & 84.84 & 73.29 & 84.96\\
\midrule
Ours
& \textbf{63.64} & \textbf{65.22} & \textbf{65.23} & \textbf{66.33} & \textbf{71.68} & \textbf{77.56} & \textbf{83.08} & \textbf{87.60} & \textbf{75.14} & \textbf{85.03} \\

\bottomrule

\end{tabular}
}
\caption{Comparison to the baselines. We measure the classification accuracy on test sets with different bias severities.
The best accuracy values are in bold. The hyphen mark `-' means it is not applicable. Results with standard deviations are provided in the Supplementary.}
\vspace{-4mm}
\label{tab:best_test}
\end{center}
\end{table*}

\begin{table}[t!]
\centering
\resizebox{1.0\linewidth}{!}{
\setlength{\tabcolsep}{0.4em}
\def\arraystretch{1.0}%
\begin{tabular}{ l | c c | c c}
\toprule 

\multicolumn{1}{c|}{\multirow{2}{*}{}}

& \multicolumn{2}{c|}{$\mathcal{D}^\text{BN}_\text{cand}$ - $\mathcal{D}^\text{BN}$}  
& \multicolumn{2}{c}{$\mathcal{D}^\text{BN}$/$\mathcal{D}$ (\%)} 

\\ \cmidrule(lr){2-3} \cmidrule(lr){4-5} 

Dataset   
& BA 
& BC
& BA
& BC 

\\
\midrule
Waterbirds 
& 26.50 \stdv{5.32} & 0.75 \stdv{0.83} & 2.75 \stdv{0.31} & 79.69 \stdv{3.72} \\

BFFHQ 
& 199.80 \stdv{40.14} & 8.00 \stdv{2.76} & 0.46 \stdv{0.09} & 50.00 \stdv{1.04} \\

BAR
& 30.60 \stdv{3.83} & 3.20 \stdv{1.60} & 3.58 \stdv{0.14} & 47.14 \stdv{5.71} \\
\bottomrule
\end{tabular}}
\caption{Effectiveness of BN score on excluding BA samples. $\mathcal{D}^\text{BN}_\text{cand}$ - $\mathcal{D}^\text{BN}$ presents the number of excluded samples when constructing $\mathcal{D}^\text{BN}$ from $\mathcal{D}^\text{BN}_\text{cand}$. $\mathcal{D}^\text{BN}$/$\mathcal{D}$ indicates that the ratio of samples in $\mathcal{D}^\text{BN}$ to the samples in $\mathcal{D}$.}

\label{tab:auxcomposition}
\end{table}

\section{Experiments}
\subsection{Experimental settings}
\label{subsec:expsetting}
\textbf{Dataset.\enskip}
We utilize Waterbirds~\cite{sagawa2019distributionally}, biased FFHQ (BFFHQ)~\cite{biaswap}, and BAR~\cite{nam2020learning} for the experiments.
Each dataset contains different types of target class and bias attributes: 
Waterbirds - \{bird type, background\}, BFFHQ - \{age, gender\}, and BAR - \{action, background\}.
The former and the latter in the bracket indicate the target class and the bias attribute, respectively. 
To be specific, 
the Waterbirds dataset has two bird classes: waterbirds and landbirds.
Most of the waterbirds are in the water background, and most of the landbirds are in the land background.
In the training dataset of BFFHQ, most young people are female while most old people are male.
The word `young' indicates an age ranging between 10 and 29, and `old' indicates an age ranging between 40 and 59.
Lastly, the BAR dataset consists of six classes of action (e.g., fishing), where the background (e.g., water surface) is highly correlated with each class.
Following the previous studies~\cite{nam2020learning, disentangled, lee2022revisiting}, we validate our model's effectiveness under different levels of bias severity, \ie, a ratio of BC samples to the total training samples: 0.5\%, 1\%, 2\%, and 5\%.
In the test sets, the spurious correlations found in the training set do not exist. 
More details are provided in Supplementary.

\noindent\textbf{Evaluation.\enskip}
We report the best accuracy of the test set averaged over five independent trials with different random seeds.
The Waterbirds dataset has an extremely skewed test dataset composed of 4,600 landbirds and 1,194 waterbirds. 
This can mislead the debiasing performance as the model may achieve high classification accuracy by simply predicting most images as landbirds. 
We measure the classification accuracy for each class and report their average value to obtain an accurate understanding of the effectiveness of methods, regardless of class frequencies.
Also, for the BFFHQ, we report the best accuracy of BC samples in the test set, following the previous works~\cite{disentangled, lee2022revisiting}.
For the analyses, we utilize the datasets with 1\% bias severity.

\subsection{Comparison to previous works}
We compare the classification accuracy on the test sets between the baselines and ours in Table~\ref{tab:best_test}.
For baselines, we employ a vanilla model trained with the CE loss, the methods using explicit bias label (\ie, LNL~\cite{kim2019LNL}, EnD~\cite{EnD}), presuming the type of the bias (\ie, HEX~\cite{wang2018hex}, ReBias~\cite{bahng2019rebias}), and assuming the bias information is unknown (\ie, LfF~\cite{nam2020learning}, DisEnt~\cite{disentangled}, LfF+BE~\cite{lee2022revisiting}, DisEnt+BE~\cite{lee2022revisiting}). 
Our approach achieves state-of-the-art performance in comparison to the previous methods including those utilizing explicit bias labels.
The results exhibit that our method improves performance robustly across various levels of bias severity, even under the constraints of extreme bias severity (\eg, 0.5 \%).
This shows that providing explicit spatial guidance for intrinsic features effectively encourages the model to learn debiased representations, leading to performance improvement.

\begin{figure}[t]
\centering
     \includegraphics[width=\linewidth]{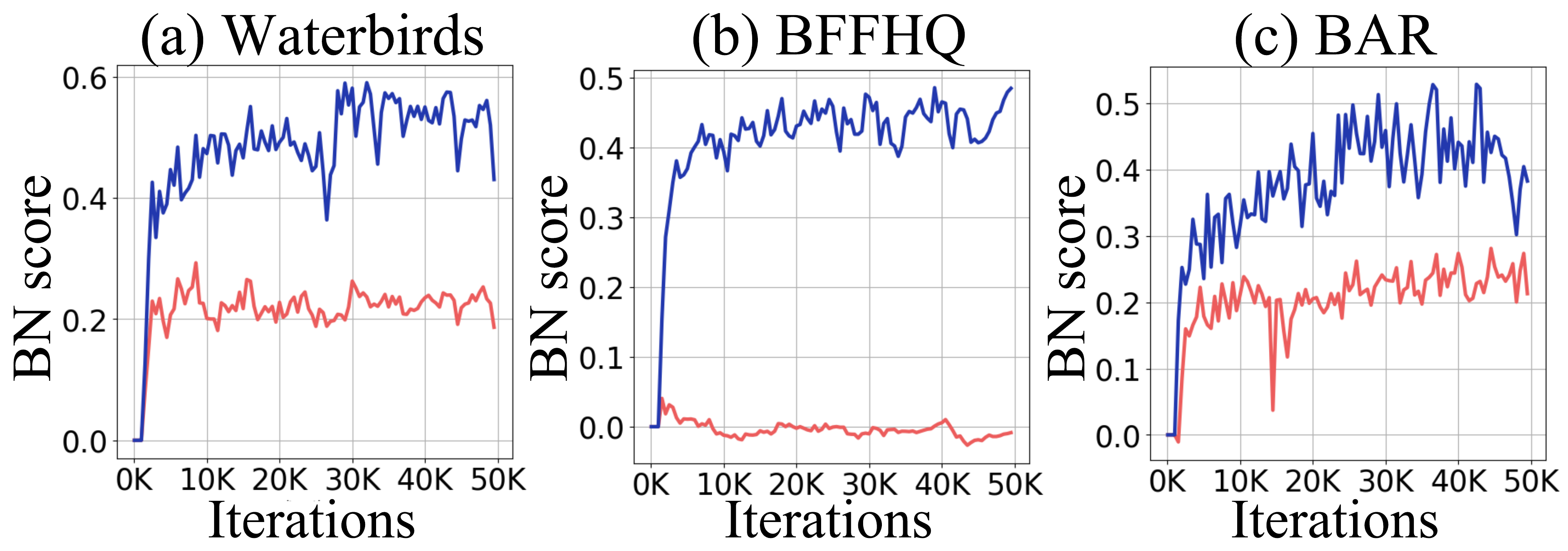}
     \caption{Visualization of BN scores of the samples in $\mathcal{D}^\text{BN}_\text{cand}$ during the training. The red lines and the blue lines indicate the BN scores of BA and BC samples, respectively.}
     \vspace{-1.5mm}
     \label{fig:score_vis}
\end{figure}

\begin{figure*}[t]
\centering
     \includegraphics[width=1.0\textwidth]{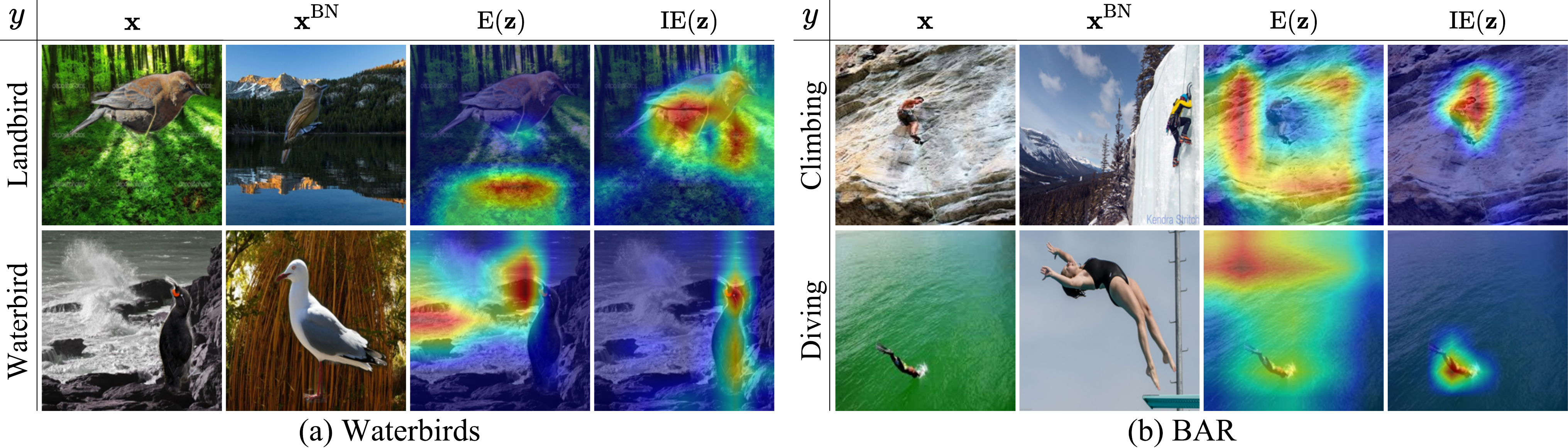}
     \caption{Visualization of the spatial guidance using (a) Waterbirds and (b) BAR dataset. Given bias-contrastive pairs, $\mathbf{x}$ and $\mathbf{x}^\text{BN}$, $\text{E}(\mathbf{z})$ indicates the regions originally focused on by $f_d$ and $\text{IE}(\mathbf{z})$ shows the regions highlighted by our IE weight.}
     \label{fig:ie_score_vis}
\end{figure*}

\begin{figure*}[t]
\centering
     \includegraphics[width=1.0\textwidth]{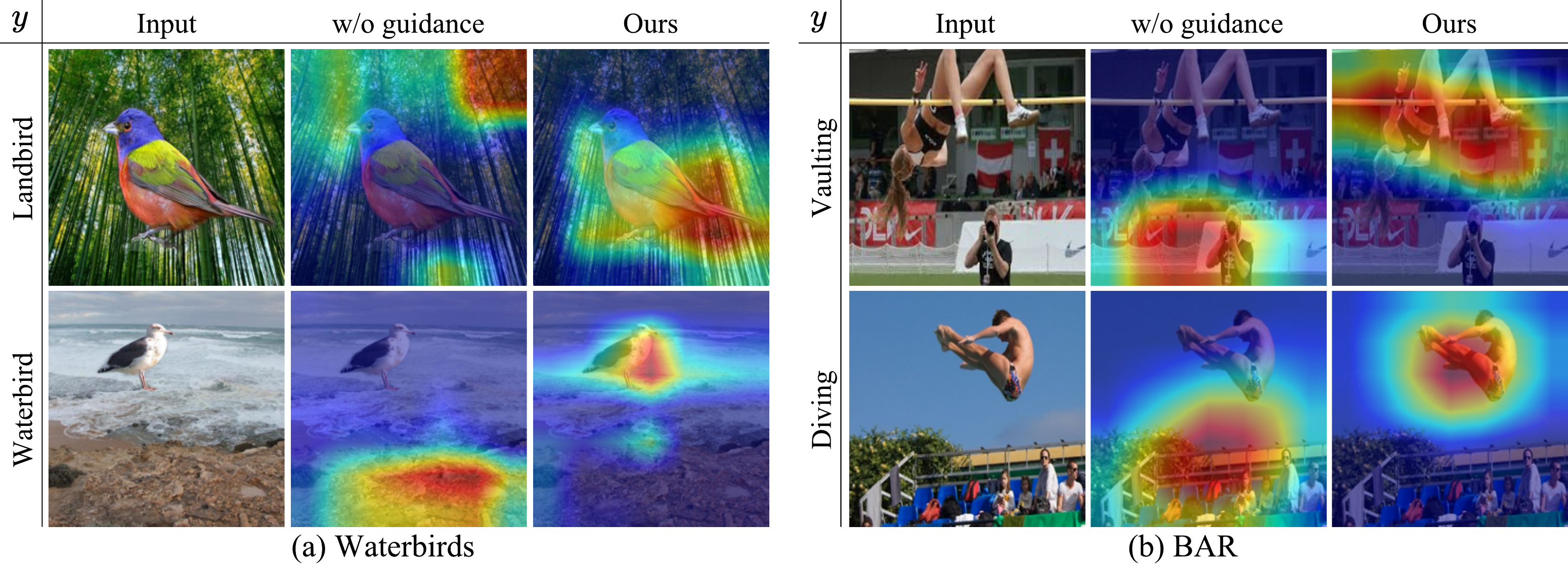}
     \caption{Comparison of the region focused by a debiased model trained with and without our method. We compare Grad-CAM results on the test set of (a) Waterbirds and (b) BAR.}
     \vspace{-2mm}
     \label{fig:cam}
\end{figure*}

\subsection{Analysis of BN score} 
\label{subsec:analysis_bn}

We analyze our BN score that identifies and emphasizes BC samples in $\mathcal{D}^\text{BN}_\text{cand}$ during training $f_d$.
In this section, we assess the effectiveness of the BN score on excluding BA samples from $\mathcal{D}^\text{BN}_\text{cand}$.
Also, we evaluate the efficacy of the BN score as a loss weight by investigating the BN scores of the samples in $\mathcal{D}^\text{BN}_\text{cand}$.

\noindent\textbf{Effectiveness of BN score on excluding BA samples.\enskip}
Table~\ref{tab:auxcomposition} presents how the BN score effectively filters out BA samples from $\mathcal{D}^\text{BN}_\text{cand}$ while preserving BC samples when constructing $\mathcal{D}^\text{BN}$.
The first two columns present the number of the BA and BC samples excluded from $\mathcal{D}^\text{BN}_\text{cand}$ to construct  $\mathcal{D}^\text{BN}$, respectively.
Also, the last two columns represent the ratio of the number of the BA and BC samples in $\mathcal{D}^\text{BN}$ to that in $\mathcal{D}$, respectively.
For the analysis, we use $\mathcal{D}^\text{BN}$ at the 50K-th iteration and report the mean value of the five independent trials.
Here, we expect $\mathcal{D}^\text{BN}$ to contain a maximal number of BC samples while including a minimal number of BA samples.
As shown in the first two columns of Table~\ref{tab:auxcomposition}, our BN score excludes a large number of BA samples while minimizing the loss of BC samples. 
As a result, $\mathcal{D}^\text{BN}$ preserves around 50-80\% of BC samples, while containing a minimal number of BA samples compared to $\mathcal{D}$, as presented in the last two columns.

\noindent\textbf{Efficacy of BN score as loss weight.\enskip}
We utilize the BN score to upweight the training loss when BC samples are chosen as $\mathbf{x}^\text{BN}$.
To verify its effectiveness as a loss weight, we compare the BN scores of BC samples and BA samples in $\mathcal{D}^\text{BN}_\text{cand}$ during the training for the Waterbirds, BFFHQ, and BAR dataset.
In Fig.~\ref{fig:score_vis}, we present the average BN scores of BA (red line) and BC samples (blue line) in $\mathcal{D}^\text{BN}_\text{cand}$ at every 500 iterations.
Since BN scores are recorded after the 1K-th iteration, the BN scores until the 1K-th iteration are reported as zero.
The BN scores of BC samples in the BFFHQ dataset mostly range from 0.4 to 0.5 while the scores of BA samples are close to 0.
This indicates that the BN score as a loss weight in the BFFHQ imposes a much larger value on the BC samples, while approximately zero values on the BA samples.
Also, the BN scores of BC samples in the Waterbirds and the BAR dataset become twice larger than the scores of BA samples.
The result shows that the BN score effectively emphasizes BC samples compared to the BA samples in $\mathcal{D}^\text{BN}_\text{cand}$ during the training.
Further analysis of the BN score is included in the Supplementary.



\subsection{Analysis of intrinsic feature guidance}
We conduct a qualitative analysis of the regions emphasized by our intrinsic feature guidance during the training and the features learned by $f_d$ after training.
We use the Waterbirds and BAR datasets with 1\% of bias severity for the analysis.


\noindent\textbf{Visualization of the guidance during training.\enskip}
In Fig.~\ref{fig:ie_score_vis}, we visualize the features emphasized by our IE weight $\text{IE}(\mathbf{z})$.
For comparison, we also visualize $\text{E}(\mathbf{z})$, the features focused by the model before applying $\text{IE}(\mathbf{z})$ for guidance.
We select a BA sample as $\mathbf{x}$ and a BC sample as $\mathbf{x}^\text{BN}$ from the training data for the analysis.
For the Waterbirds dataset in Fig.~\ref{fig:ie_score_vis}(a), $\text{IE}(\mathbf{z})$ highlights the wings or the beak of the bird compared to $\text{E}(\mathbf{z})$, where the forest or the water (\ie, bias attributes) is highlighted.
Also, in Fig.~\ref{fig:ie_score_vis}(b), $\text{E}(\mathbf{z})$ focuses more on the bias attributes such as rocks or the water than the intrinsic attributes.
In contrast, $\text{IE}(\mathbf{z})$ emphasizes the action of the human that is less exploited compared to the bias features in $\text{E}(\mathbf{z})$.
The results demonstrate that our guidance successfully identifies and enhances under-exploited intrinsic features during the training.

\noindent \textbf{Effect of intrinsic feature guidance on debiasing.\enskip}
We qualitatively evaluate the effectiveness of the intrinsic feature guidance by investigating the visual explanation maps of the test samples.
We compare the Grad-CAM~\cite{gradcam} results of the model trained with and without our method in Fig~\ref{fig:cam}.
The Grad-CAM results highlight the features that the model employs to predict the input as its ground-truth label.
In Fig.~\ref{fig:cam} (a), while the model trained without guidance focuses on the forest or the sea, ours focuses on the tail or a curved shape of the bird's body.
Additionally, Fig.~\ref{fig:cam}~(b) shows that ours focuses on the motion of the human rather than the backgrounds that are concentrated on by the model trained without our guidance.
The results verify that our method successfully encourages the model to learn intrinsic features from the training dataset, improving the robustness of the model against dataset bias.

\subsection{Ablation study}
\label{subsec:ablation}
As shown in Table~\ref{tab:ablation}, we perform ablation studies to verify the effectiveness of the individual components in our method.
The results of ours are reported in the last row.

\noindent\textbf{Importance of $\mathbf{x}^\text{BN}$ selection and BN score as loss weight.\enskip}
We demonstrate the efficacy of adopting BC samples as $\mathbf{x}^\text{BN}$.
We train the model by sampling $\mathbf{x}^\text{BN}$ from three different datasets: $\mathcal{D}$, $\mathcal{D}^\text{BN}_\text{cand}$, and $\mathcal{D}^\text{BN}$.
In the first row of Table~\ref{tab:ablation}, we randomly sample $\mathbf{x}^\text{BN}$ from the training dataset $\mathcal{D}$ without using the BN score.
In the second row, we train the model with $\mathbf{x}^\text{BN}$ sampled from $\mathcal{D}^\text{BN}_\text{cand}$, where BA samples are roughly filtered out using the early-stopped biased models.
The model in the third row is trained with $\mathbf{x}^\text{BN}$ from $\mathcal{D}^\text{BN}$ that mainly includes BC samples using our BN score.
The results show a gradual improvement in debiasing performance as more BC samples are selected as $\mathbf{x}^\text{BN}$.
This is because auxiliary samples without bias attributes prevent the common features from including the bias features, composing a bias-contrastive pair with the input. 
Therefore, our guidance effectively enhances the intrinsic features during the training.
Finally, the last row in Table~\ref{tab:ablation} presents that employing the BN score $s(\mathbf{x}^\text{BN})$ to reweight the losses further enhances performance by emphasizing the usage of BC samples as $\mathbf{x}^\text{BN}$.

\noindent\textbf{Training objectives.\enskip}
We examine the impact of each training objective, $\mathcal{L}_\text{guide}$ and $\mathcal{L}_\text{BN}$, in our method.
We report the performance of the model trained without $\mathcal{L}_\text{guide}$ (the fourth row) and without $\mathcal{L}_\text{BN}$ (the fifth row) in Table~\ref{tab:ablation}. 
The model trained without $\mathcal{L}_\text{guide}$ exhibits degraded performance, facing difficulties in identifying where to focus to learn intrinsic features. 
Similarly, training the model without $\mathcal{L}_\text{BN}$ also results in a performance decrease.
The results verify that $\mathcal{L}_\text{BN}$ successfully supports the IE weight to identify intrinsic features among the common features by learning class-discerning features from $\mathbf{x}^\text{BN}$.
The model that incorporates both $\mathcal{L}_\text{guide}$ and $\mathcal{L}_\text{BN}$ demonstrates the best performance (the last row of Table~\ref{tab:ablation}).

\begin{table}[t!]
\centering
\resizebox{1.0\linewidth}{!}{
\setlength{\tabcolsep}{0.3em}
\def\arraystretch{1.3}%
\begin{tabular}{ c c | c c | c c c}
\toprule 

$\mathcal{L}_\text{guide}$ & $\mathcal{L}_\text{BN}$ & $\mathbf{x
}^\text{BN}$ & \begin{tabular}{@{}c@{}}$s(\mathbf{x}^\text{BN})$ as  \\ loss weight \end{tabular}  & Waterbirds & BFFHQ & BAR \\
\midrule
\textcolor{black}{\boldcheckmark} 
& \textcolor{black}{\boldcheckmark} 
&  $\mathcal{D}$    
& \textcolor{black}{\boldxmark} 
& 62.79 \stdv{1.21} & 71.04 \stdv{2.55} & 73.36 \stdv{1.40} \\
\textcolor{black}{\boldcheckmark} 
& \textcolor{black}{\boldcheckmark} 
&  $\mathcal{D}^\text{BN}_\text{cand}$   
& \textcolor{black}{\boldxmark} 
& 64.65 \stdv{1.23} & 75.64 \stdv{1.87} & 74.27 \stdv{0.66} \\
\textcolor{black}{\boldcheckmark} 
& \textcolor{black}{\boldcheckmark} 
&  $\mathcal{D}^\text{BN}$    
& \textcolor{black}{\boldxmark} 
& 65.10 \stdv{0.87} & 77.08 \stdv{2.05} & 74.62 \stdv{1.07} \\

\midrule
\textcolor{black}{\boldxmark} 
& \textcolor{black}{\boldcheckmark} 
&  $\mathcal{D}^\text{BN}$     
& \textcolor{black}{\boldcheckmark} 
& 63.81 \stdv{1.24} & 76.92 \stdv{1.03} & 74.03 \stdv{1.13}\\
\textcolor{black}{\boldcheckmark} 
& \textcolor{black}{\boldxmark} 
&  $\mathcal{D}^\text{BN}$      
& \textcolor{black}{\boldcheckmark} 
& 62.10 \stdv{3.35} & 74.84 \stdv{2.00} & 74.87 \stdv{1.51} \\

\midrule
\textcolor{black}{\boldcheckmark} 
& \textcolor{black}{\boldcheckmark} 
&  $\mathcal{D}^\text{BN}$    
& \textcolor{black}{\boldcheckmark} 
& \textbf{65.22} \stdv{0.95} &  \textbf{77.56} \stdv{1.24} & \textbf{75.14} \stdv{0.82} \\

\bottomrule
\end{tabular}}
\caption{Ablation study on the proposed training objectives, the dataset that $\mathbf{x}^\text{BN}$ is sampled from, and the BN score of $\mathbf{x}^\text{BN}$ as a loss weight.
The check mark (\textcolor{black}{\boldcheckmark}) denotes the inclusion of the corresponding method, while the cross mark (\textcolor{black}{\boldxmark}) indicates the exclusion of the component in the experiment.}
\vspace{-2mm}
\label{tab:ablation}
\end{table}

\section{Conclusion}
In this paper, we propose a debiasing method that explicitly provides the model with spatial guidance for intrinsic features.
Leveraging an auxiliary sample, we first identify intrinsic features by investigating the class-discerning features commonly appearing in a bias-contrastive pair.
Our IE weight enhances the intrinsic features that have not been focused on yet in the input by a debiased model.
To construct the bias-contrastive pair without bias labels, we introduce a bias-negative (BN) score that tracks the classification loss of a biased model to distinguish BC samples from BA samples during the training.
The effectiveness of our method is demonstrated through experiments on synthetic and real-world datasets with varying levels of bias severity. 
We believe this work sheds light on the significance of providing explicit guidance on the intrinsic attributes for debiasing.

\section*{Acknowledgments}
This work was supported by the Institute for Information \& communications Technology Promotion(IITP) grant funded by the Korea government(MSIT) (No.2019-0-00075, Artificial Intelligence Graduate School Program(KAIST), the National Research Foundation of Korea (NRF) grant funded by the Korea government (MSIT) (No. NRF-2022R1A2B5B02001913 \& No. 2022R1A5A7083908) and partly by Kakao Corporation. 

\clearpage
{
    \small
    \bibliographystyle{ieeenat_fullname}
    \bibliography{main}
}

\appendix

\clearpage
\setcounter{page}{1}
\maketitlesupplementary

%

This supplementary material offers further analysis of our approach, additional experimental results, the details of the datasets and implementation, limitations, and future work. 
\cref{supp:lossBN} and \cref{supp:negBN} provide the analysis of the bias-negative (BN) score as a loss weight and samples with negative BN score, respectively.
\cref{supp:bc_sample_selection} analyzes the effect of BC samples in $\mathcal{D}^\text{BN}$ on debiasing performance. Also, \cref{supp:recent_sample_selection} compares the recent sample selection methods with ours.
Moreover, \cref{supp:qual_result} and \cref{supp:quantitative_results} present additional qualitative results regarding the guidance and additional quantitative results, respectively.
\cref{supp:data} and \cref{supp:implementdetail} provide the details about the dataset and implementation.
Lastly, \cref{supp:limit} discusses the limitations and future work.

\section{Additional analysis of the BN score as a loss weight}
\label{supp:lossBN}
\begin{figure}[h]
\centering
     \includegraphics[width=\linewidth]{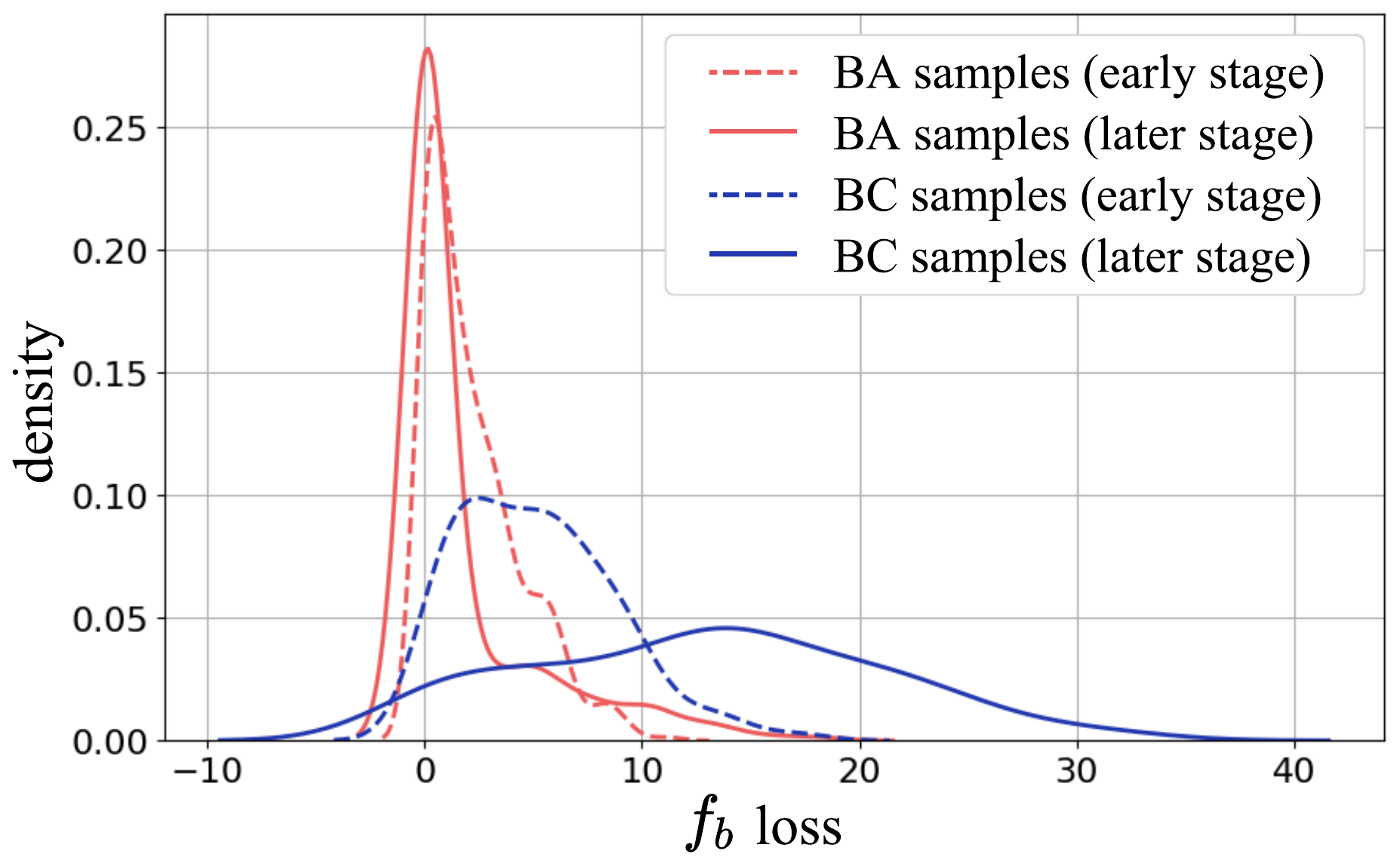}
     \caption{The distributions of $f_b$'s classification loss of samples in $\mathcal{D}^\text{BN}_\text{cand}$. The red and blue lines denote the losses of BA and BC samples, respectively. The dotted and solid lines indicate the losses at the early and later stages of the training, respectively. Best viewed in color.}
     \label{fig:loss_compare}
\end{figure}

As described in Sec.~3.4 in the main paper, we utilize the BN score of $\mathbf{x}^\text{BN}$ (\ie, $s(\mathbf{x}^\text{BN})$) to reweight the guidance loss $\mathcal{L}_\text{guide\_sim}$ and the BN loss $\mathcal{L}_\text{BN}$.
The BN score as a loss weight is designed to upweight the losses when bias-conflicting (BC) samples are selected as $\mathbf{x}^\text{BN}$, which further encourages our IE weight to enhance the intrinsic features.
For verification, we present that the BN score has a much larger value on the BC samples compared to bias-aligned (BA) samples during the training in Fig.~2 in the main paper.

Since $s(\mathbf{x}^\text{BN})$ has a larger value when the current $f_b$ loss of $\mathbf{x}^\text{BN}$ is larger than that of the early stage of training, the results imply that the $f_b$ loss of BC samples largely increases as training proceeds compared to BA samples.

To further verify this, we present $f_b$'s classification loss of samples in $\mathcal{D}^\text{BN}_\text{cand}$ during the training in Fig.~\ref{fig:loss_compare}.
The BFFHQ dataset~\cite{biaswap} with a bias severity of 1\% is used for the analysis.
In Fig.~\ref{fig:loss_compare}, the dotted lines denote the distribution of $f_b$'s classification loss at the early stage of training (1K-th iteration), and the solid lines indicate that of the later stage of training (50K-th iteration).
The results show that the $f_b$ loss of BC samples (blue lines) largely increases at the later stage of training compared to the early stage, unlike BA samples (red lines).
This demonstrates that the BN score as a loss weight can effectively upweight the training losses when BC samples are chosen as $\mathbf{x}^\text{BN}$.

\section{Samples having negative BN score}
\label{supp:negBN}
\begin{figure}[h]
\centering
     \includegraphics[width=\linewidth]{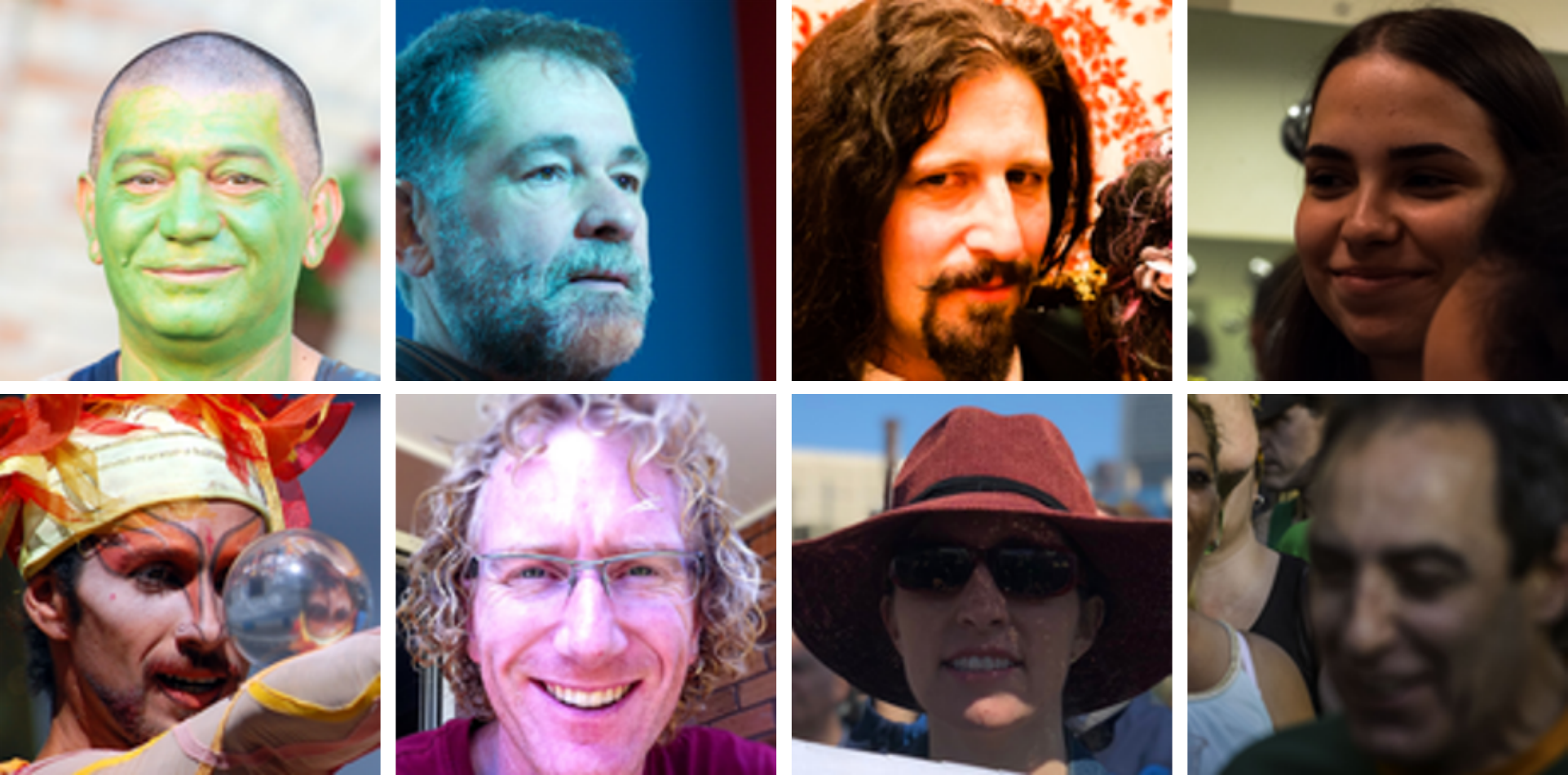}
     \caption{The examples of samples that have negative BN scores at the later stage of training.}
     \label{fig:supp_negativeBN}
\end{figure}

\begin{figure*}[t]
    \centering
    \includegraphics[width=1.0\textwidth]{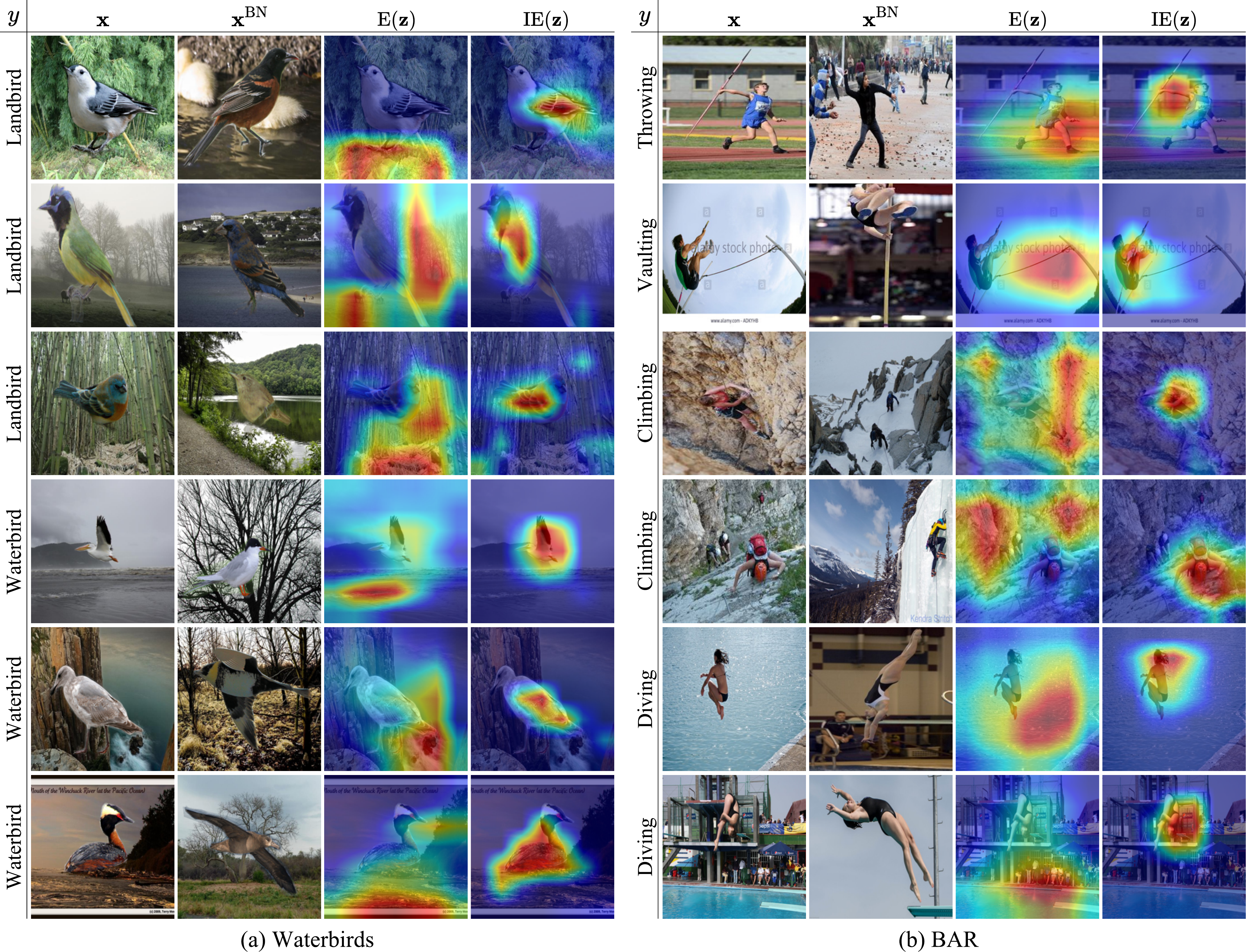}
    \caption{
    Additional visualization results of the spatial guidance using (a) Waterbirds and (b) BAR dataset. Given bias-contrastive pairs, $\mathbf{x}$ and $\mathbf{x}^\text{BN}$, $\text{E}(\mathbf{z})$ indicates the regions originally focused on by $f_d$ and $\text{IE}(\mathbf{z})$ shows the regions highlighted by our IE weight.}
    \label{fig:supp_ie_score}
\end{figure*}

As mentioned in Sec.~3.2 in the main paper, we further filter out the samples with negative BN scores from $\mathcal{D}_\text{cand}^\text{BN}$ to mainly exploit the BC samples as $\mathbf{x}^\text{BN}$.
Here, we expect that the samples with negative BN scores are mostly BA samples.
To investigate the samples with negative BN scores, we chose the samples that were erroneously incorporated into $\mathcal{D}_\text{cand}^\text{BN}$ initially but excluded at the later stage of training (\ie, 50K-th iteration), exhibiting negative BN scores.
This process is repeated five times, and we visualize the samples chosen more than three times in Fig.~\ref{fig:supp_negativeBN}.
We use the BFFHQ dataset with a 1\% bias severity for the experiment.

We observe that the samples with negative BN scores in $\mathcal{D}_\text{cand}^\text{BN}$ are mostly BA samples.
As shown in the figure, while the samples obviously contain bias attributes (\ie, features representing female or male), the samples mostly have extreme shade, blur, saturation, or unusual makeup, exhibiting non-typical appearance.
Although the bias attributes are known to be easy to learn, the non-typical appearance prevents $f_b$ from detecting such bias attributes in the early stage of training.
In Sec.~4.5 of the main paper, we verify that employing such BA samples as $\mathbf{x}^\text{BN}$ largely degrades the debiasing performance by allowing the bias attributes to be included in the common features between $\mathbf{x}$ and $\mathbf{x}^\text{BN}$.
Our BN score effectively alleviates this issue by filtering out such BA samples from $\mathcal{D}_\text{cand}^\text{BN}$.

\begin{figure*}[t]
    \centering
    \includegraphics[width=1.0\textwidth]{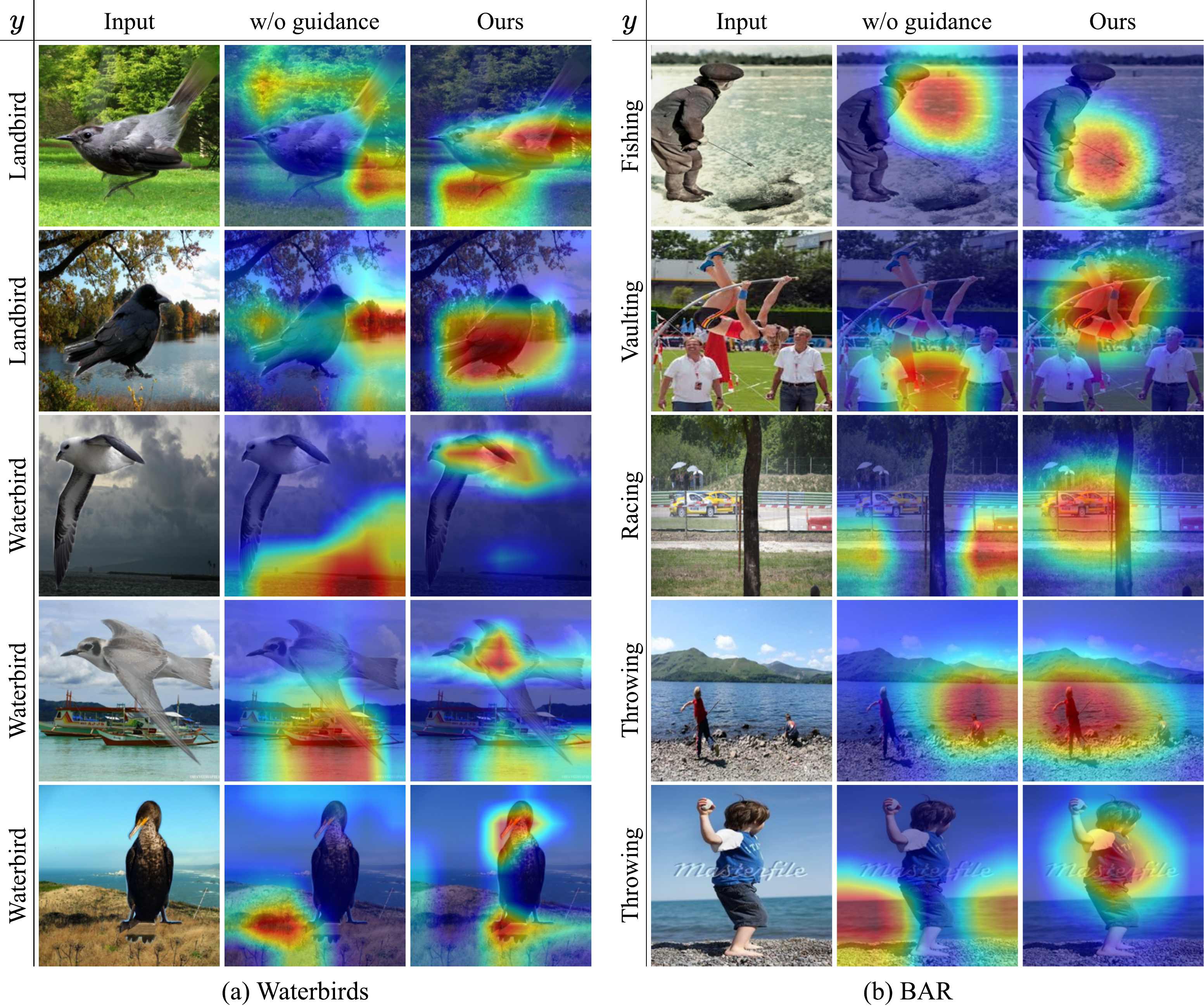}
    \caption{Additional comparison of the region focused by a debiased model trained with and without our method. We compare Grad-CAM results on the test set of (a) Waterbirds and (b) BAR.}
    \label{fig:supp_gradcam}
\end{figure*}

\section{Importance of BN sample selection}
\label{supp:bc_sample_selection}
We analyze the effect of the BC sample ratio in $\mathcal{D}^\text{BN}$ on debiasing performance.
We measure the accuracy using the BFFHQ dataset with a 1\% bias severity by varying the number of BA and BC samples in $\mathcal{D}^\text{BN}$.
Table~\ref{tab:bn_quality} shows that higher accuracy is achieved for more BC samples and a lower ratio of BA to BC samples in $\mathcal{D}^\text{BN}$.
Overall, our method constantly shows performance gain, except for the last column (\{${\text{\#BC in }\mathcal{D}^\text{BN}}/{\text{\#BC in }\mathcal{D}}$, ${\text{\#BA in }\mathcal{D}^\text{BN}}/{\text{\#BC in }\mathcal{D}^\text{BN}}$\}-\{1.0, 10.0\}).
It is crucial not to select too many BA samples as $\mathbf{x}^\text{BN}$. 

\begin{table}[t!]
\centering
\resizebox{\linewidth}{!}{
\setlength{\tabcolsep}{0.25em}
\def\arraystretch{1.1}%
\begin{tabular}{ l | c c c | c c c c}
\toprule

${\text{\#BC in }\mathcal{D}^\text{BN}}/{\text{\#BC in }\mathcal{D}}$
& 0.1
& 0.5
& 1.0
& 1.0
& 1.0
& 1.0
& 1.0
\\

${\text{\#BA in }\mathcal{D}^\text{BN}}/{\text{\#BC in }\mathcal{D}^\text{BN}}$
& 0.0
& 0.0
& 0.0
& 0.1
& 1.0
& 2.0
& 10.0

\\
\midrule

Accuracy 
& 75.84 & 78.12 & 81.40 
& 80.24 & 77.48 &  75.48 & 70.90\\

\bottomrule
\end{tabular}}
\caption{Importance of $\mathbf{x}^\text{BN}$ selection. 
}
\label{tab:bn_quality}
\end{table}

\section{Comparison to recent sample selection methods}
\label{supp:recent_sample_selection}
Our BN score is designed to further filter out BA samples in $\mathcal{D}^\text{BN}_\text{cand}$, improving debiasing performance (Sec.~4.5 in the main paper).
We compare BC sample selection in recent methods with ours using BFFHQ with a 1\% bias severity.
Let $\mathcal{S}$ be a set of samples identified as BC samples from training data $\mathcal{D}$.
$\{\text{\#BC in } \mathcal{S} / \text{\#BC in } \mathcal{D} , \text{\#BA in }\mathcal{S} / \text{\#BC in }\mathcal{S}\}$ is \{75.63, 10.18\}-BE~\cite{lee2022revisiting}, \{27.29, 4.32\}-DCWP~\cite{dcwp2023park}, and \{50.0, 0.89\}-Ours, respectively.
Our method has the least number of BA compared to BC samples in $\mathcal{S}$ while preserving half of the total BC samples.

\section{Additional qualitative results}
\label{supp:qual_result}
\subsection{Visualization of the guidance during training}

In addition to Fig.~3 of the main paper, we provide supplementary qualitative results that present the features that the current model $f_d$ focuses on (\ie, $\text{E}(\mathbf{z})$) and the features emphasized by the guidance (\ie, $\text{IE}(\mathbf{z})$) during the training in Fig.~\ref{fig:supp_ie_score}.
We use the Waterbirds and BAR datasets with a bias severity of 1\% for the analysis.
We train $f_d$ during 10K iterations and obtain the visual explanation map $\text{E}(\mathbf{z})$ for the ground-truth label using Grad-CAM~\cite{gradcam}.
The min-max normalization is applied to the values of $\text{E}(\mathbf{z})$ and $\text{IE}(\mathbf{z})$ for visualization. 
We intentionally select the BA sample and BC sample as $\mathbf{x}$ and $\mathbf{x}^\text{BN}$, respectively, to compose a bias-negative pair.

As shown in Fig.~\ref{fig:supp_ie_score}, our IE weight (\ie, $\text{IE}(\mathbf{z})$) appropriately emphasizes the regions of the intrinsic features while $\text{E}(\mathbf{z})$ shows that the current model $f_d$ relies on the bias attributes for prediction.
For example, in the Waterbirds dataset, $\text{IE}(\mathbf{z})$ properly enhances the intrinsic features of a bird such as wings, a body, or a neck, while $\text{E}(\mathbf{z})$ highlights the background features such as the land, the forest or the water.
Also, in the BAR dataset, $\text{IE}(\mathbf{z})$ emphasizes the arm throwing the javelin, the motion of a person vaulting, climbing, or diving, while the current $f_d$ mainly focuses on the biased features such as the playing field, the sky, the mountain, or the water.
These results verify the validity of our IE weight $\text{IE}(\mathbf{z})$ as guidance for emphasizing the intrinsic features in $\mathbf{x}$ that are under-exploited yet.

\subsection{Effect of intrinsic feature guidance on debiasing}

We present an additional qualitative analysis regarding the effectiveness of the intrinsic feature guidance to supplement Fig.~4 in the main paper.
Fig.~\ref{fig:supp_gradcam} illustrates the Grad-CAM~\cite{gradcam} results of the model trained with and without our method.
Here, the model trained without our method is the same as LfF+BE~\cite{lee2022revisiting}.
We train the models with the Waterbids and the BAR datasets with a bias severity of 1\% and apply the Grad-CAM to the test samples for visualization.
The highlighted regions indicate the features that the model mainly employs for prediction.

Fig.~\ref{fig:supp_gradcam}~(a) shows that our approach properly focuses on the intrinsic features of the bird (\eg, wings, a beak, or feet), while the model trained without our guidance mostly concentrates on the bias features (\eg, the water or trees).
For the BAR dataset in Fig.~\ref{fig:supp_gradcam}~(b), our model principally exploits the action of a person (\eg, fishing, vaulting, or throwing) or the racing car for prediction, while the model without our guidance focuses on the backgrounds (\eg, the playing field or the water). 
The results demonstrate the effectiveness of our method in guiding the model to learn intrinsic features.

\begin{table*}[h!]
\centering
\resizebox{0.75\linewidth}{!}{
\setlength{\tabcolsep}{1.2em}
\def\arraystretch{1.0}%
\begin{tabular}{l|cccc}
\toprule 
\multicolumn{1}{c}
{\multirow{2}{*}{Method}} &
 \multicolumn{4}{c}{Waterbirds} 
\\ \cmidrule(lr){2-5}
\multicolumn{1}{c}{}&
 0.5 & 1.0 & 2.0 & 5.0 
\\
\midrule







Vanilla~\cite{He2015resnet} & 57.41 \stdv{0.74} & 58.07 \stdv{1.00} & 61.04 \stdv{0.55} & 64.13 \stdv{0.14} \\
HEX~\cite{wang2018hex} & 57.88 \stdv{0.83} & 58.28 \stdv{0.67} & 61.02 \stdv{0.48} & 64.32 \stdv{0.62} \\
LNL~\cite{kim2019LNL} & 58.49 \stdv{0.81} & 59.68 \stdv{0.78} & 62.27 \stdv{0.91} & 66.07 \stdv{1.15} \\
EnD~\cite{EnD} & 58.47 \stdv{0.97} & 57.81 \stdv{1.04} & 61.26 \stdv{0.54} & 64.11 \stdv{0.52} \\
ReBias~\cite{bahng2019rebias} & 55.44 \stdv{0.24} & 55.93 \stdv{0.66} & 58.53 \stdv{0.52} & 62.14 \stdv{1.03} \\
LfF~\cite{nam2020learning} & 60.66 \stdv{0.77} & 61.78 \stdv{1.53} & 58.92 \stdv{2.93} & 61.43 \stdv{1.92} \\
DisEnt~\cite{disentangled} & 59.59 \stdv{1.67} & 60.05 \stdv{0.82} & 59.76 \stdv{1.26} & 64.01 \stdv{1.36} \\
LfF+BE~\cite{lee2022revisiting} & 61.22 \stdv{2.54} & 62.58 \stdv{1.12} & 63.00 \stdv{1.18} & 63.48 \stdv{0.56} \\
DisEnt+BE~\cite{lee2022revisiting} & 51.65 \stdv{1.45} & 54.10 \stdv{1.04} & 53.43 \stdv{1.42} & 54.21 \stdv{1.36} \\
\midrule
Ours & \textbf{63.64} \stdv{1.63} & \textbf{65.22} \stdv{0.95} & \textbf{65.23} \stdv{1.06} & \textbf{66.33} \stdv{1.42} \\

\bottomrule

\end{tabular}
}
\caption{Comparison to the baselines. We measure the classification accuracy on test sets of the Waterbirds dataset with different bias severities.
The best accuracy values are in bold. Results with standard deviations are provided in the Supplementary.}
\label{tab:supp_main_synthetic}
\end{table*}

\begin{table*}[h]
\centering
\resizebox{1.0\linewidth}{!}{
\setlength{\tabcolsep}{1.2em}
\def\arraystretch{1.0}%
\begin{tabular}{l|cccc|cc}
\toprule 
\multicolumn{1}{c}
{\multirow{2}{*}{Method}} &
\multicolumn{4}{c}{BFFHQ} & \multicolumn{2}{c}{BAR}
\\ \cmidrule(lr){2-5}\cmidrule(lr){6-7}
\multicolumn{1}{c}{}&
0.5 & 1.0 & 2.0 & 5.0 & 1.0 & 5.0 
\\
\midrule


Vanilla~\cite{He2015resnet} & 55.64 \stdv{0.44} & 60.96 \stdv{1.00} & 69.00 \stdv{0.50} & 82.88 \stdv{0.49} & 70.55 \stdv{0.87} & 82.53 \stdv{1.08} \\
HEX~\cite{wang2018hex} & 56.96 \stdv{0.62} & 62.32 \stdv{1.21} & 70.72 \stdv{0.89} & 83.40 \stdv{0.34} & 70.48 \stdv{1.74} & 81.20 \stdv{0.68} \\
LNL~\cite{kim2019LNL} & 56.88 \stdv{1.13} & 62.64 \stdv{0.99} & 69.80 \stdv{1.03} & 83.08 \stdv{0.93} & - & - \\
EnD~\cite{EnD} & 55.96 \stdv{0.91} & 60.88 \stdv{1.17} & 69.72 \stdv{1.14} & 82.88 \stdv{0.74} & - & - \\
ReBias~\cite{bahng2019rebias} & 55.76 \stdv{1.50} & 60.68 \stdv{1.24} & 69.60 \stdv{1.33} & 82.64 \stdv{0.64} & 73.04 \stdv{1.04} & 83.90 \stdv{0.82} \\
LfF~\cite{nam2020learning} & 65.19 \stdv{3.23} & 69.24 \stdv{2.07} & 73.08 \stdv{2.70} & 79.80 \stdv{1.09} & 70.16 \stdv{0.77} & 82.95 \stdv{0.27} \\
DisEnt~\cite{disentangled} & 62.08 \stdv{3.89} & 66.00 \stdv{1.33} & 69.92 \stdv{2.72} & 80.68 \stdv{0.25} & 70.33 \stdv{0.19} & 83.13 \stdv{0.46} \\
LfF+BE~\cite{lee2022revisiting} & 67.36 \stdv{3.10} & 75.08 \stdv{2.29} & 80.32 \stdv{2.07} & 85.48 \stdv{2.88} & 73.36 \stdv{0.97} & 83.87 \stdv{0.82} \\
DisEnt+BE~\cite{lee2022revisiting} & 67.56 \stdv{2.11} & 73.48 \stdv{2.12} & 79.48 \stdv{1.80} & 84.84 \stdv{2.11} & 73.29 \stdv{0.41} & 84.96 \stdv{0.69} \\
DCWP~\cite{dcwp2023park}&64.08 \stdv{1.08} &67.44 \stdv{2.87} &75.24 \stdv{1.73} &85.00 \stdv{0.94} &69.63 \stdv{0.85} &81.89 \stdv{0.68} \\ 
\midrule
Ours & \textbf{71.68} \stdv{1.74} & \textbf{77.56} \stdv{1.24} & \textbf{83.08} \stdv{1.69} & \textbf{87.60} \stdv{1.68} & \textbf{75.14} \stdv{0.82} & \textbf{85.03} \stdv{0.64} \\

\bottomrule

\end{tabular}
}
\caption{Comparison to the baselines. We measure the classification accuracy on test sets of the BFFHQ and BAR datasets with different bias severities.
The best accuracy values are in bold. The hyphen mark (-) means it is not applicable. Results with standard deviations are provided in the Supplementary.}
\label{tab:supp_main_real}
\end{table*}


\section{Additional quantitative results}
\label{supp:quantitative_results}
\subsection{Quantitative results with standard deviations}
\label{supp:resultwstd}
In Table~1 of our main paper, we report the quantitative comparison results with classification accuracies on the test set which are averaged across five independent experiments with different random seeds.
We additionally provide the standard deviations of the classification accuracies in Table~\ref{tab:supp_main_synthetic} and Table~\ref{tab:supp_main_real}.
Each table shows the results of the synthetic dataset (\ie, Waterbirds) and the real-world dataset (\ie, BFFHQ and BAR), respectively.
Since the BAR dataset lacks explicit bias labels, approaches such as LNL and EnD that necessitate explicit bias labels are not applicable to the BAR dataset.
The baseline results for the BFFHQ and the BAR dataset are from the results reported in BE~\cite{lee2022revisiting} except for DCWP~\cite{dcwp2023park}.


\subsection{Comparison to recent baseline}
Our primary contribution lies in providing the model with explicit spatial guidance for intrinsic features by examining features that commonly appear in bias-contrastive pairs.
The intrinsic feature exists in generally appearing features within a class, however, this property has not been tackled to provide intrinsic feature guidance in prior studies to the best of our knowledge.
While recent debiasing approaches aim to encourage the model to learn intrinsic features, they fail to directly indicate where the model should focus to learn the features.

For instance, MaskTune~\cite{masktune2022} expects the model to learn intrinsic features by fine-tuning the model with the data whose already-explored area is masked out using GradCAM.
However, simply exploring the unmasked area cannot inform the model where exactly the intrinsic features are located.
In this case, the model may rather focus on non-intrinsic features during the fine-tuning.
We experiment on real-world datasets with a 1\% bias severity: \{58.00, 69.42\}-MaskTune and \{77.56, 75.14\}-Ours for \{BFFHQ, BAR\}.
Ours achieves better debiasing performance by providing explicit spatial guidance for intrinsic features based on common features in bias-contrastive pairs.

A recent pair-wise debiasing method $\mathcal{X}^2$-model~\cite{Zhang_2023_CVPR} encourages the model to retain intra-class compactness using samples generated via feature-level interpolation between BC and BA samples.
However, $\mathcal{X}^2$-model does not inform the model where the intrinsic features are located in the interpolated features.
Simply making samples closer to the interpolated samples does not assure the model to focus on the intrinsic features.
In contrast, our method directly encourages the model to focus on the area of the intrinsic features.

Also, we conduct a quantitative comparison to the recently proposed debiasing approach, DCWP~\cite{dcwp2023park}, in Table~\ref{tab:supp_main_real}.
We use real-world datasets, BFFHQ and BAR, with various bias severity.
For a fair comparison, we utilize ResNet18, which is the same architecture as ours.
The ImageNet pretrained weight is employed only for the BAR dataset.
The results demonstrate the superiority of our method over the DCWP, where ours provides the model with explicit guidance for intrinsic features for debiasing, unlike DCWP.

\subsection{Worst accuracy between the accuracy of BA and BC samples in Waterbirds}

\begin{table*}
\begin{center}

\resizebox{1.0\linewidth}{!}{
\setlength{\tabcolsep}{0.4em}
\def\arraystretch{1.1}%
\begin{tabular}{c|cccccccccc}
\toprule 
BS
& Vanilla~\cite{He2015resnet}
& HEX~\cite{wang2018hex}
& LNL~\cite{kim2019LNL}
& EnD~\cite{EnD}
& ReBias~\cite{bahng2019rebias}
& LfF~\cite{nam2020learning}
& DisEnt~\cite{disentangled}
& LfF+BE~\cite{lee2022revisiting}
& DisEnt+BE~\cite{lee2022revisiting}
& Ours
\\

\midrule
{0.5}
& 24.08 \stdv{1.56} & 28.20 \stdv{3.07} & 26.08 \stdv{1.64} & 28.29 \stdv{3.53} & 27.00 \stdv{1.10} & 56.22 \stdv{6.07} & 38.07 \stdv{11.01} & 55.15 \stdv{2.78} & 36.60 \stdv{10.88} & 59.12 \stdv{3.67}  \\

\midrule
{1.0}
& 24.78 \stdv{2.45} & 26.32 \stdv{2.90} & 29.72 \stdv{3.45} & 25.69 \stdv{2.41} & 27.95 \stdv{1.56} & 59.07 \stdv{3.40} & 47.02 \stdv{7.26} & 55.53 \stdv{1.60} & 28.35 \stdv{4.17} & 63.05 \stdv{1.97}  \\

\midrule
{2.0}
& 34.39 \stdv{2.24} & 32.12 \stdv{2.89} & 33.92 \stdv{1.94} & 32.94 \stdv{1.48} & 32.16 \stdv{0.76} & 53.07 \stdv{6.74} & 44.93 \stdv{8.54} & 52.91 \stdv{2.62} & 31.08 \stdv{6.01} & 61.71 \stdv{4.94}  \\

\midrule
{5.0}
& 38.34 \stdv{1.05} & 39.08 \stdv{0.92} & 43.22 \stdv{1.94} & 40.91 \stdv{1.11} & 39.72 \stdv{1.11} & 58.05 \stdv{2.37} & 52.96 \stdv{6.33} & 48.48 \stdv{3.72} & 37.92 \stdv{6.47} & 58.60 \stdv{3.32}  \\

\bottomrule

\end{tabular}
}
\caption{The worst accuracy between the accuracy of BA and BC samples in the Waterbirds dataset. BS is bias severity.}
\label{tab:wb_babc_accs}
\end{center}
\end{table*}
To further analyze our model's performance on the Waterbirds dataset, we measure the accuracy of BA and BC samples separately, where the class accuracy values are averaged.
Then, we report the worst accuracy between them in Table~\ref{tab:wb_babc_accs}.
The results show that ours achieves the highest worst accuracy compared to other baselines.

\section{Detailed description of datasets}
\label{supp:data}

\begin{figure*}[t]
    \centering
    \includegraphics[width=1.0\textwidth]{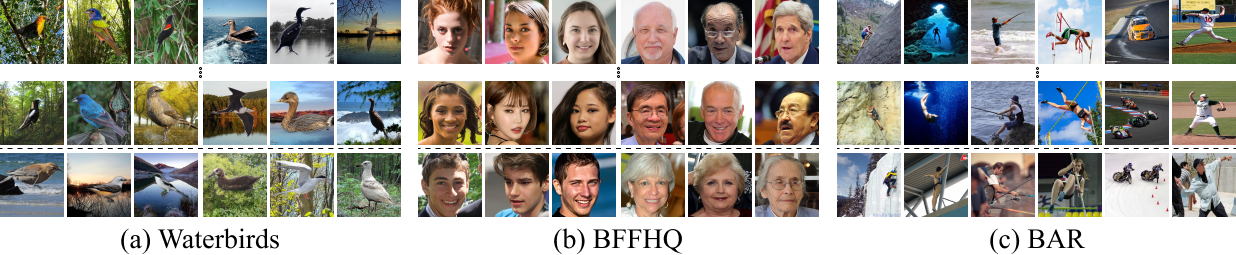}
    \caption{Visualization of datasets used in the experiments. A group of three columns represents each class for (a) Waterbirds-\{Landbird, Waterbird\} and (b) BFFHQ-\{Young, Old\}, and each column of (c) BAR-\{Climbing, Diving, Fishing, Vaulting, Racing, Throwing\} represents a distinct class. The samples above the dashed line are bias-aligned samples and the below ones are bias-conflicting samples.}
    \label{fig:dataset}
\end{figure*}

We utilize Waterbirds~\cite{sagawa2019distributionally}, BFFHQ~\cite{biaswap}, and BAR~\cite{nam2020learning} dataset.
First, the Waterbirds dataset is composed of two classes of bird images and has background bias.
In the training set, the waterbirds are mostly with the water background and the landbirds are with the land background.
The number of BA samples and that of BC samples are balanced in the test set.
By following Sagawa \textit{et al.}~\cite{sagawa2019distributionally}, we utilize two datsets, the Caltech-UCSD Birds-200-2011 (CUB) dataset~\cite{WahCUB_200_2011} and the Places\footnote{\label{4}CC BY} dataset~\cite{zhou2017places}, to construct the Waterbirds dataset.
The bird images are segmented from the CUB dataset, and the segmented birds are combined with the background images from the Place dataset.
We employ the code released by Sagawa \textit{et al.}~\cite{sagawa2019distributionally}\footnote{\label{5}https://github.com/kohpangwei/group\_DRO} for constructing the dataset. 
As mentioned in the repository, a few landbirds (Eastern Towhees, Western Meadowlarks, and Western Wood Pewees) in the original dataset are incorrectly labeled as waterbirds.
Therefore, we correct their labels to landbirds for the experiments.

The BFFHQ is initially presented by Kim~\textit{et al.}~\cite{biaswap} and constructed by modifying the FFHQ dataset \footnote{\label{3}BY-NC-SA 4.0}.
In the BFFHQ, the bias attribute is the gender and the intrinsic attribute is the age.
Specifically, most of the young people are female, and most of the old people are male in the training dataset.

Lastly, the BAR dataset is introduced by Nam~\textit{et al.}~\cite{nam2020learning}.
The dataset contains six action classes (\ie, Climbing, Diving, Fishing, Vaulting, Racing, Throwing) and each class is biased to a certain background (\ie, RockWall, Underwater, WaterSurface, Sky, APavedTrack, PlayingField).
In the test set, such correlations do not exist.
For the experiments, we use the BFFHQ dataset and BAR dataset released by Lee \textit{et al.}~\cite{lee2022revisiting}\footnote{\label{2}https://github.com/kakaoenterprise/BiasEnsemble}. 
The examples of the BA samples and BC samples in each dataset are shown in Fig.~\ref{fig:dataset}.

\section{Implementation details}
\label{supp:implementdetail}
Following the previous studies~\cite{nam2020learning, disentangled, lee2022revisiting}, we utilize ResNet18~\cite{He2015resnet} for the biased model $f_b$ and the debiased model $f_d$. 
Also, $f_d^\text{emb}$ indicates the subnetwork before the average pooling layer, and $f_d^\text{cls}$ consists of an average pooling layer and a linear classifier that outputs logits, where $f_d(\mathbf{x}) = f_d^{\text{cls}}\left(f_d^{\text{emb}}(\mathbf{x})\right)$.
Before training, $f_b$ and $f_d$ are initialized with the ImageNet pretrained weight for the BAR dataset, while we randomly initialize the models for the other datasets.
This is because the size of the BAR dataset is extremely small compared to the others~\cite{lee2022revisiting}.

During training $f_d$, we employ the sample reweighting value $w(\mathbf{x})$ termed as relative difficulty score~\cite{nam2020learning}, as mentioned in Sec.~3.4 in the main paper.
$w(\mathbf{x})$ is calculated as follows:
\begin{equation}
    w(\mathbf{x})=\frac{\mathcal{L}_\text{CE}(f_b(\mathbf{x}), y)}{\mathcal{L}_\text{CE}(f_b(\mathbf{x}), y) + \mathcal{L}_\text{CE}(f_d(\mathbf{x}), y)}.
\end{equation}
This score assigns a high weight to the BC samples and a low weight to the BA samples.
This encourages $f_d$ to mainly learn intrinsic features by emphasizing BC samples with $w(\mathbf{x})$.

The models are trained for 50K iterations with a batch size of 64.
The horizontal flip and a random crop with a size of 224 are used for data augmentation during the training. 
All the models are trained with the Adam optimizer.
The learning rate is set as 1e-4 for the Waterbirds and the BFFHQ dataset, and 1e-5 for the BAR dataset. 
The hyperparameters of $\alpha_l$, $\alpha_s,$ and $\tau$ are set as $0.1$, $0.9$, and $2$, respectively, for all the datasets.
We apply class-wise max normalization to our BN score to consider the different ranges of the scores across the classes for stability of training.

During the training, we aim to select an auxiliary sample that has no bias attribute but has the same class label with $\mathbf{x}$ as $\mathbf{x}^\text{BN}$ from $\mathcal{D}^\text{BN}$. 
If no sample in $\mathcal{D}^\text{BN}$ has the same label as $\mathbf{x}$, we select the sample that has the same label with $\mathbf{x}$ from $\mathcal{D}^\text{BN}_\text{cand}$.
In a case where there's no sample with the same label as $\mathbf{x}$ in both $\mathcal{D}^\text{BN}$ and $\mathcal{D}^\text{BN}_\text{cand}$, we sample $\mathbf{x}^\text{BN}$ that has the same label with $\mathbf{x}$ from $\mathcal{D}$.

As described in Sec.~3.1 in the main paper, we utilize the pretrained biased models to construct $\mathcal{D}^\text{A}$, following the previous work~\cite{lee2022revisiting}.
We utilize ResNet18~\cite{He2015resnet} for the pretrained biased models, and all the pretrained biased models are randomly initialized before training.
The models are trained for 1K iterations with the generalized cross entropy (GCE) loss~\cite{zhang2018generalized}.
Within each model, the samples with a high ground-truth probability (\textit{i.e.,} $\ge$ 0.99) are considered as BA samples.
Based on majority voting, we collect the samples that are considered as the BA sample by the majority of the models and construct the bias-amplified dataset $\mathcal{D}^\text{A}$.
We use five pretrained biased models following the study of Lee \etal~\cite{lee2022revisiting}.
Lee \etal demonstrate that adopting the additional biased models requires a negligible amount of additional computational costs and memory space.
Note that the same biased models are utilized when constructing $\mathcal{D}_\text{cand}^\text{BN}$.

\section{Limitations and future work}
\label{supp:limit}
Although our BN score effectively encourages BC samples to be mainly adopted as auxiliary inputs, the auxiliary inputs still can include a few BA samples, as shown in Table~2 of Sec.~4.3 in the main paper.
Accordingly, such BA samples may interfere with the model to capture the intrinsic features.
Identifying intrinsic attributes without relying on auxiliary inputs can be one promising future work.

In addition, since our IE weight is designed to enhance intrinsic features by imposing spatially different values on the features, our method might be more effective especially when bias attributes are located in different regions from the intrinsic attributes.
In this regard, applying channel-wise re-weighting~\cite{hu2018senet} to our approach will further improve the general applicability of our method.

Despite the limitations above, we believe that our work poses the importance of enhancing intrinsic attributes for debiasing.


\end{document}